\documentclass[review,11pt]{ReportTemplate}
\usepackage{hyperref}
\usepackage{url}
\usepackage{eso-pic}
\usepackage{amsmath, amssymb, amsthm, bm}
\usepackage{mathtools}
\usepackage{algorithm}
\usepackage{float}
\usepackage{algorithmic}
\usepackage{graphicx}
\usepackage{subfig}
\usepackage{booktabs}
\usepackage{tikz}
\usepackage{xcolor}
\usepackage{pifont} 
\usetikzlibrary{arrows.meta,positioning,shapes.geometric}
\usepackage{listings}
\usepackage[framemethod=tikz]{mdframed} 
\newmdenv[
  backgroundcolor=gray!3,
  linecolor=gray!50,
  roundcorner=3pt,
  linewidth=0.8pt,
  innerleftmargin=8pt,
  innerrightmargin=8pt,
  innertopmargin=6pt,
  innerbottommargin=6pt,
  skipabove=8pt,
  skipbelow=8pt
]{examplebox}

\newcommand{\cmark}{\textcolor{green!60!black}{\ding{51}}}
\newcommand{\xmark}{\textcolor{red!70!black}{\ding{55}}}

\theoremstyle{plain}
\newtheorem{theorem}{Theorem}[section]
\newtheorem{proposition}[theorem]{Proposition}

\newtheorem{corollary}[theorem]{Corollary}
\theoremstyle{definition}
\newtheorem{definition}[theorem]{Definition}

\theoremstyle{remark}

\newcommand{\E}{\mathbb{E}}
\newcommand{\Prb}{\mathbb{P}}

\newcommand{\grad}{\nabla_{\!\theta}}

\makeatletter
\def\els@aparagraph[#1]#2{\elsparagraph[{#1}]{#2.}}
\AtBeginDocument{\ClearShipoutPicture} 
\makeatother

\usepackage{xurl}
\DeclareRobustCommand{\rev}[1]{\textcolor{black}{#1}}

\begin{document}

\begin{frontmatter}	
	\title{Reinforcement Learning with Verifiable yet Noisy Rewards under Imperfect Verifiers}
	\author{
\begin{tabular}{@{}l@{}}
\textbf{Xin-Qiang Cai}$^1$        \quad
\textbf{Wei Wang}$^{2,1}$ \quad 
\textbf{Feng Liu}$^{2}$ \quad
\textbf{Tongliang liu}$^{3}$ \quad \\
\quad \quad \quad \quad \quad \quad \quad \textbf{Gang Niu}$^{1}$ \quad
\textbf{Masashi Sugiyama}$^{1,2}$
\end{tabular}
\vspace{0mm} \\
$^{1}$RIKEN AIP, Tokyo, Japan \quad $^{2}$The University of Tokyo, Tokyo, Japan \\ $^{3}$The University of Melbourne, Melbourne, Australia \\ $^{4}$The University of Sydney, Sydney, Australia
}

\begin{abstract}
Reinforcement Learning with Verifiable Rewards (RLVR) replaces costly human labeling with automated verifiers. To reduce verifier hacking, many RLVR systems binarize rewards to $\{0,1\}$, but imperfect verifiers inevitably introduce \emph{false negatives} (FN, rejecting correct answers) and \emph{false positives} (FP, accepting incorrect ones). We formalize verifier unreliability as a stochastic reward channel with asymmetric noise rates $\rho_0$ and $\rho_1$---the FP rate and the FN rate, respectively. From this abstraction we derive two lightweight corrections: (i) a \emph{backward} correction that yields an unbiased surrogate reward and thus an unbiased policy-gradient estimator in expectation, and (ii) a \emph{forward} correction that reweights score-function terms so the expected update aligns with the clean gradient direction and requires only the FN rate. We implement both as lightweight hooks in a group relative policy optimization pipeline, both corrections improve RLVR reasoning under synthetic and real verifier noise, with the forward variant being more stable under heavier noise. We further use a lightweight appeals mechanism to estimate FN rates online and show that the correction can complement stronger verifier-side baselines.
\end{abstract}

\end{frontmatter}

\section{Introduction}
\label{sec:introduction}
Reinforcement Learning with Verifiable Rewards (RLVR) offers a scalable paradigm for improving the reasoning abilities of Large Language Models (LLMs) by replacing expensive human annotation with automated feedback~\citep{RLVR2025}. In this problem, a policy is trained using rewards from a verifier that automatically checks the correctness of a model's output~\citep{deepseekmath}. The efficacy of this approach, however, heavily depends on the verifier's reliability. Emerging evidence reveals that verifiers are systematically fallible in two critical ways: they can accept incorrect solutions~(\emph{false positives; FPs}) or reject correct ones~(\emph{false negatives; FNs})~\citep{xu2025tinyv,zhao2025onetoken}.

\textit{FPs} have been widely documented as a vulnerability of LLM-based verifiers. Recent studies showed that LLM judges can be swayed by superficial cues---e.g., popular specialized verifiers, such as GPT-4o, give \textit{35 \% -- 66.8\%} FP rate when the answer starts with \textit{Let’s solve this problem step by step}~\citep{zhao2025onetoken,shi2025positionbias,shi2024judgedeceiver,chen2024humansjudge}. Conversely, \textit{FNs} are common with rule-based verifiers. These checkers, while highly precise, are often brittle; they may reject valid solutions that are formatted differently, expressed in an algebraically equivalent form, or embedded in explanatory text~\citep{math-verify}. A recent analysis of a math-RL dataset found that over \textit{38\%} of responses flagged as incorrect by a rule-based system were in fact correct, a gap that a lightweight LLM verifier could partially close~\citep{xu2025tinyv}. Both FPs and FNs materially degrade RLVR training: FNs deprive the agent of informative gradients and slow convergence, while FPs reward hackable patterns and inflate returns during policy optimization~\citep{xu2025tinyv,huang2025pitfalls}.

\begin{figure*}[t]
  \centering
  \includegraphics[width=0.9\linewidth]{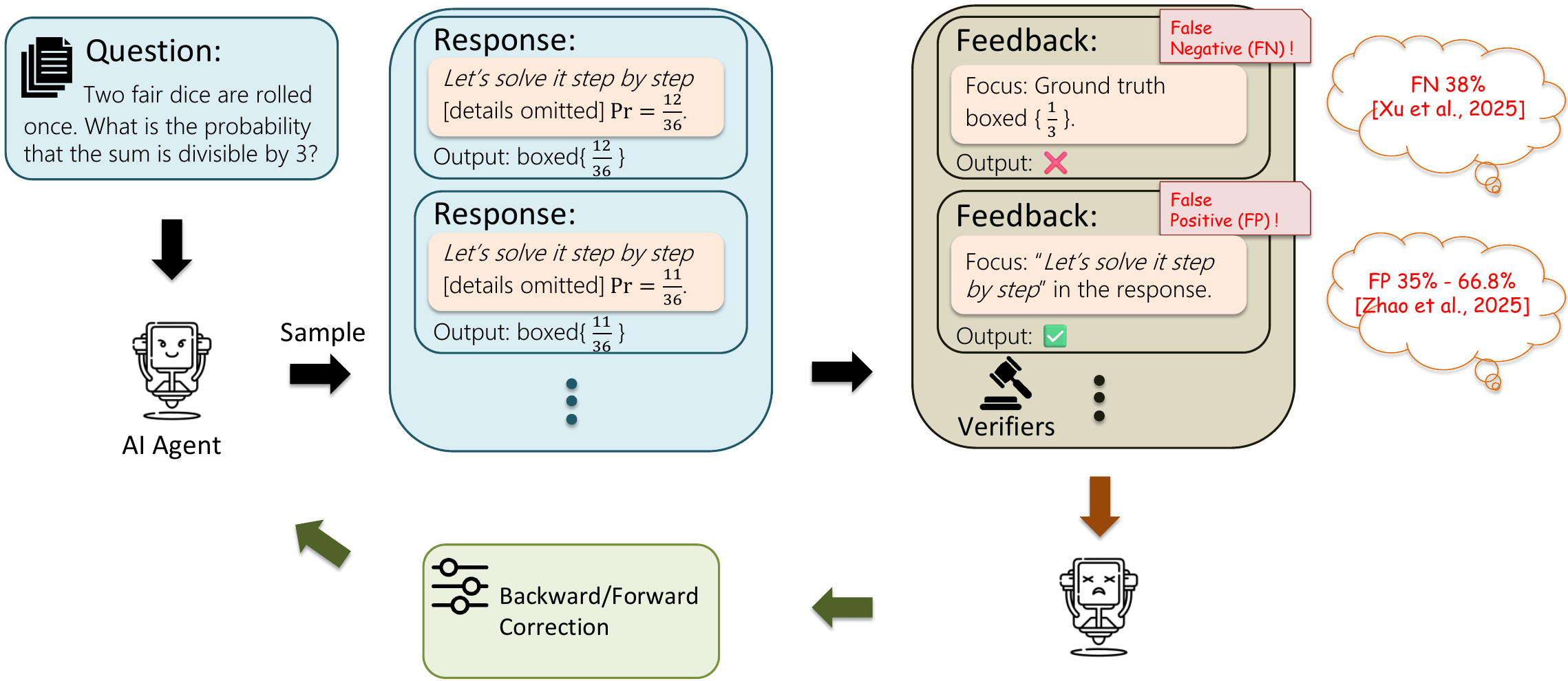}%
  \caption{Verifier-noise flow in RLVR. An AI agent produces candidate solutions that are scored by automated verifiers. While verifiers would yield \textit{false negatives} ($\frac{12}{36}$ vs.\ $\frac{1}{3}$, reaching $38\%$ rates~\citep{xu2025tinyv}) and \textit{false positives} (mislead by ``Let's solve it step by step...'', reaching $35\% - 66.8\%$ rates~\citep{zhao2025onetoken}), confusing the agent; applying our backward/forward corrections restores correct signals.}
  \vspace{-4mm}
  \label{fig:noise-flow}
\end{figure*}


Motivated by these failures, we address these challenges by explicitly \emph{treating verifier errors as noise} in the reward signal. We model the verifier as a stochastic reward channel that corrupts the true, latent reward with verifier-conditional probabilities---the FP rate and the FN rate. Building on this formulation, we derive two estimators to counteract this noise. Our first method, \emph{noisy Policy Gradient with Backward Correction (PGBC)}, inverts the noise process to compute an unbiased estimator of the true reward, which can be used as a drop-in replacement in any advantage-estimation in RL. Our second method, \emph{noisy Policy Gradient with Forward Correction (PGFC)}, directly reweights the terms of the policy gradient to ensure its expected direction aligns with the clean gradient. This approach requires only an estimate of the FN rate to achieve directional correctness, making it suitable for the often harder-to-estimate FP rate~\citep{xu2025tinyv}. The pipeline is illustrated in Figure~\ref{fig:noise-flow}.

We integrate these corrections into a group relative policy optimization (GRPO)~\citep{deepseekmath}-based RLVR pipeline and demonstrate their effectiveness on competitive math reasoning benchmarks~\citep{deepseekmath}. Both methods consistently outperform uncorrected training and achieve performance nearly equivalent to that of noise-free scenarios, with the forward correction offering the fastest and most stable convergence. We also propose a practical online estimation scheme for the FN rate: a querying mechanism where a low-cost LLM verifier like TinyV~\citep{xu2025tinyv} re-evaluates outputs rejected by a primary rule-based checker. This hybrid approach provides a reliable estimate of the FN rate with minimal computational overhead, recovering near-oracle performance. Furthermore, we stress-test our approach under complex, non-i.i.d.\ noise distributions that violate our FP and FN modeling, but both PGBC and PGFC corrections remain robust and effective even in these more challenging conditions. Our contributions are: (i) a formal verifier-channel model for RLVR that captures the prevalent FP and FN errors; (ii) backward and forward corrections derived from RL principles to calibrate policy learning; and (iii) a practical implementation for online noise estimation that makes these corrections viable in real-world systems.

\section{Related Works}
\label{sec:related}

\paragraph{Reasoning with LLMs.}
A large body of work improves LLM reasoning through prompting and search at inference time. Chain-of-Thought (CoT) prompting elicits step-by-step rationales and boosts arithmetic and commonsense reasoning~\citep{Wei2022CoT}, while self-consistency aggregates diverse reasoning paths to further improve robustness~\citep{Wang2022SelfConsistency}. Decomposition-based prompting, such as Least-to-Most, solves complex problems via ordered sub-problems~\citep{Zhou2022LeastToMost}; search-based schemes like Tree-of-Thoughts explicitly explore and evaluate alternative reasoning branches~\citep{Yao2023ToT}. Orthogonally, training-time verifier signals can be used to re-rank candidate solutions, as in \citet{Cobbe2021GSM8K}. Our work is complementary: rather than proposing a new inference-time strategy or re-ranking scheme, we address \emph{how} to perform \emph{policy optimization} when the reward itself—supplied by a verifier—is noisy.

\vspace{-2mm}
\paragraph{RLVR and verifier reliability.}
Recent math-RL pipelines combine on-policy RL (e.g., GRPO) with programmatic or LLM-based verifiers to yield verifiable rewards~\citep{mroueh2025grpo}. However, LLM-as-a-judge is prone to systematic biases, including position bias and prompt-based attacks, producing \emph{FPs} (accepting incorrect solutions)~\citep{LLMJudgeSurvey2024,Thakur2024JudgingJudges,shi2025positionbias,zhao2025onetoken,shi2024judgedeceiver}. Conversely, exact-match or brittle parsers in rule-based checkers can miss algebraic equivalences and text-formatted answers, leading to widespread \emph{FNs}~\citep{Li2025VerifyBench,huang2025pitfalls}. A recent study showed that over 38\% of model responses labeled incorrect by a rule-based pipeline were in fact correct, and introduced a lightweight verifier (TinyV) that recovers many such cases with minimal overhead~\citep{xu2025tinyv}. Prior efforts therefore improved the \emph{verifier} (e.g., better judges or appeals) or evaluation suites; in contrast, we treat verifier errors as verifier-conditional noise and directly \emph{correct the policy gradient} so that learning remains aligned with the clean objective even when the verifier is imperfect.

\vspace{-4mm}
\paragraph{Learning with noisy labels.}
While our method is derived from RL, it shares intuition with the literature on learning with label noise. \citet{wang2020reinforcement} modeled reward corruption via a class-conditional confusion matrix and derived an unbiased surrogate reward for Q-learning algorithms to solve control tasks. In supervised learning, a central thread is to make empirical-risk minimization robust either by \emph{correcting the loss} given a noise-transition model or by \emph{avoiding} or \emph{down-weighting} suspected noisy examples~\citep{song2020survey,ICML:Li+etal:2021}. The former includes the unbiased-risk estimators of \citet{natarajan2013} and the now-standard loss corrections of \citet{patrini2017}. A second family avoids the noise explicitly by relying on the ``small-loss first'' memorization dynamics of deep nets: mentor methods (MentorNet) learn a weighting network that feeds cleaner samples to the student, and \emph{Co-teaching}~\citep{han2018coteaching} trains two peers that exchange their selected small-loss examples; semi-supervised hybrids such as \emph{DivideMix}~\citep{li2020dividemix} split data into clean/noisy partitions via mixture modeling and co-train with consistency regularization~\citep{jiang2018mentornet}. Unlike them, we target the \emph{policy gradient estimator} in RLVR: we instantiate both backward and forward corrections inside the policy gradient estimator to stabilize and align RL under verifier noise, instead of a direct supervised-loss substitution. 

\section{Problem Setup and Algorithms}
\label{sec:problem_setup}

In this section, we introduce the problem formulations and our proposed algorithms.

\subsection{Problem Setup}
We consider the standard RLVR setting where a stochastic policy $\pi_\theta(y|x)$ generates a response $y$ for a given prompt $x$, where $\theta$ denotes learnable parameters. The goal is to maximize an objective based on the true, or \emph{clean}, reward $R^*(x,y) \in \{0,1\}$, which indicates whether the response $y$ is genuinely correct. This objective is typically regularized by a KL-divergence term to a reference policy $\pi_{\mathrm{ref}}$ to maintain stability:
\begin{equation}
  \max_\theta\; J_\beta(\theta)
  \;= \; \E_{x}\,\E_{y\sim\pi_\theta(\cdot\mid x)}\!\Big[\,R^*(x,y) - \beta\,\mathrm{KL}(\pi_\theta(\cdot\mid x)\,\|\,\pi_{\mathrm{ref}}(\cdot\mid x))\,\Big],
\end{equation}
in which $\mathrm{KL}$ denotes the Kullback-Leibler (KL)-divergence. The policy gradient with respect to the unregularized reward is given by the REINFORCE estimator~\citep{sutton1999policy}, $\nabla_\theta J(\theta) = \E[R^*(x,y)\,\sum_{t} G_t]$, where $G_t = \grad \log \pi_\theta(y_t\mid x,y_{<t})$ is the score function for the token at step $t$, and $y_{<t}$ denotes the sequence of tokens generated prior to step 
$t$. In the following, we will denote $R(x,y)$ with $R$ for brevity. In practice, the clean reward $R^*$ is unavailable. Instead, the agent receives a noisy reward $\tilde{R}(x,y) \in \{0,1\}$ from an automated verifier. We model this verifier as a stochastic \emph{reward channel} that flips the latent clean reward to an observed noisy one. This channel is characterized by asymmetric, verifier-conditional noise rates.
\begin{definition}[Verifier Reward Channel]\label{def:vrc}
  The observed verifier reward $\tilde{R}$ is generated from the latent clean reward $R^*$ according to the following conditional probabilities:
  \begin{equation}
    \Prb\bigl(\tilde{R}=1 \mid R^*=0\bigr) = \rho_0,
    \Prb\bigl(\tilde{R}=0 \mid R^*=1\bigr) = \rho_1,
  \end{equation}
  where $\rho_0, \rho_1 \in [0, 1)$ and $\rho_0 + \rho_1 < 1$.
\end{definition}
The condition $\rho_0+\rho_1<1$ implies that the verifier channel is informative and invertible. Equivalently, the $2\times2$ binary transition matrix has determinant $1-\rho_0-\rho_1>0$.

\vspace{-2mm}
\paragraph{Assumptions and scope of the verifier channel.}
Definition~\ref{def:vrc} is a deliberately lightweight abstraction of verifier unreliability.
It assumes (i) \emph{instance-independent} class-conditional noise rates $(\rho_0,\rho_1)$ that do not vary with $(x,y)$, and (ii) \emph{conditional independence}: given the latent correctness $R^*(x,y)$, the verifier's stochastic flip $\tilde R$ is independent of the policy's sampling randomness (hence independent of the score-function terms used by policy gradients).
These assumptions are standard in class-conditional noise models and allow closed-form reward/gradient corrections.
In real RLVR deployments, verifier errors can be content-dependent (e.g., spurious phrases can increase FPs), which violates (i) and can partially violate (ii).
We therefore view $(\rho_0,\rho_1)$ as \emph{effective} noise rates, and empirically test robustness to mis-specification in Section~\ref{subsec:robustness};
we further estimate $\rho_1$ online via a low-cost appeals mechanism in Section~\ref{subsec:real-world-exp}.
When rates are instance-dependent, the same formulas are treated as correcting an effective marginal channel rather than eliminating all bias; Appendix~\ref{app:impl-grpo} gives the resulting covariance residual.

Based on this noise model, we establish a linear relationship between the expected noisy reward and the clean reward, which is presented in Proposition~\ref{prop:affine-reward}.

\begin{proposition}[Connection between Corrupted Rewards and True Rewards]
  \label{prop:affine-reward}
  Under the Verifier Reward Channel model, the expectation of the noisy reward $\tilde R$ conditioned on the clean reward $R^*$ is an affine transformation of $R^*$:
  \[
    \mathbb{E}[\tilde R \mid R^*]\;=\;(1-\rho_0-\rho_1)\,R^*+\rho_0.
  \]
\end{proposition}
The proof is provided in Appendix~\ref{app:proof_affine-reward}. The central challenge of RLVR is that naively optimizing with the noisy reward $\tilde{R}$ leads to a biased policy gradient, causing the policy to learn from mistakes of the verifier. Our goal is to develop policy gradient estimators that use only the observable noisy reward $\tilde{R}$ but the expectation is either identical or parallel to the true policy gradient $\nabla_\theta J(\theta)$. We integrate these estimators within REINFORCE-style policy gradient algorithms widely used for reasoning tasks that compute normalized advantages over a group of sampled trajectories~\citep{deepseekmath}.

\subsection{Noisy Policy Gradient with Backward Correction (PGBC)}
\label{subsec:algo_backward}

Our first proposed approach aims to construct an unbiased estimator of the true reward $R^*$ by ``inverting'' the noise process. Since the expected noisy reward $\E[\tilde{R} \mid R^*]$ is an affine transformation of the clean reward $R^*$ in Proposition~\ref{prop:affine-reward}, we can solve for $R^*$ to derive a corrected reward estimator, $\widehat{R}$, that is unbiased in expectation.

\begin{theorem}[Oracle-Rate Unbiased Reward Estimator for the Binary Verifier Channel]
  \label{thm:unbiased_reward_estimator}
  Given the verifier channel with known noise rates $\rho_0$ and $\rho_1$ with $\rho_0 + \rho_1 < 1$, the estimator
  \begin{equation}
    \widehat{R} = \frac{\tilde{R} - \rho_0}{\,1 - \rho_0 - \rho_1\,}
  \end{equation}
  is an unbiased estimator of the true reward $R^*(x, y)$, i.e., $\E[\widehat{R}] = R^*$.
\end{theorem}
The proof is provided in Appendix~\ref{app:proof_thm_unbiased_asym}. The theorem is an oracle-rate population statement: unbiasedness holds when the channel rates used in the correction equal the true effective rates and before finite-sample GRPO normalization, clipping, or KL terms are applied. Theorem~\ref{thm:unbiased_reward_estimator} shows that, by replacing the noisy reward $\tilde{R}$ with $\widehat{R}$ in any standard policy gradient formulation, we obtain an unbiased estimate of the true policy gradient. This allows $\widehat{R}$ to serve as a drop-in replacement for the reward signal in complex RL algorithms, seamlessly integrating with advantage estimation and other machinery. 
The details of the algorithm are summarized in Algorithm~\ref{alg:backward}. After rolling out a set of trajectories and obtaining their rewards, we use the corrected rewards to compute the policy gradients for model update.

\begin{algorithm}[t]
  \caption{Noisy Policy Gradient with Backward Correction (PGBC)}
  \label{alg:backward}
  \begin{algorithmic}
    \STATE \textbf{Input:} Initial policy $\theta_0$; learning rate $\eta$; batch size $M$; estimates of noise rates, $(\hat\rho_0,\hat\rho_1)$.
    \LOOP
    \STATE Roll out $M$ trajectories $\{(x_i, y_i)\}_{i=1}^M$, obtaining observed rewards $\{\tilde{R}_i\}_{i=1}^M$.
    \STATE For each trajectory $i$, compute the unbiased reward estimate:
    \[
      \widehat{R}_i \leftarrow \frac{\tilde{R}_i - \hat\rho_0}{1 - \hat\rho_0 - \hat\rho_1}.
    \]
    \STATE Compute the policy gradient using the corrected rewards $\{\widehat{R}_i\}_{i=1}^M$.
    \STATE Update parameters: $\theta \leftarrow \theta + \eta\,\Delta\theta$.
    \ENDLOOP
  \end{algorithmic}
\end{algorithm}

\subsection{Noisy Policy Gradient with Forward Correction (PGFC)}
\label{subsec:algo_forward}
While PGBC provides an elegant unbiased estimator, it faces two practical challenges. First, the denominator $(1 - \rho_0 - \rho_1)$ can be small if the total noise rate is high, leading to high variance in the reward estimate $\widehat{R}$ and potentially unstable training. Second, it requires accurate estimation of both the false positive rate $\rho_0$ and the false negative rate $\rho_1$, which are usually difficult to obtain in practice. The PGFC approach is introduced to mitigate these issues.

Instead of correcting the reward itself, this method directly modifies the policy gradient estimator to ensure its expectation is correctly aligned. The core idea is to reweight the score function $G_t$ under the $t$-th token based on the observed noisy reward $\tilde{R}$. We define an update term $h_t = w_{\tilde{R}} G_t$ and choose the \emph{forward weights} $w_0$ (for $\tilde{R}=0$) and $w_1$ (for $\tilde{R}=1$) such that the expected update $\E[h_t]$ is parallel to the clean gradient $\nabla_\theta J(\theta)$. The key insight is to choose the forward weights so that the conditional expectations given the true reward $R^*$ have a specific structure.
\begin{proposition}[Conditional Expectation of Forward Weights]
  \label{prop:forward_weights_expectation}
  Let the forward weights be defined as $w_0 = \rho_1 - 1$ and $w_1 = \rho_1$. Under the Verifier Reward Channel model, the conditional expectations of the corresponding weight $w_{\tilde{R}}$ given the true reward $R^*$ are:
  \begin{enumerate}
    \item $\E[w_{\tilde{R}} \mid R^*=1] \;=\; 0$.
    \item $\E[w_{\tilde{R}} \mid R^*=0] \;=\; -(1 - \rho_0 - \rho_1)$.
  \end{enumerate}
\end{proposition}
The proof can be found in Appendix~\ref{app:proof_prop_forward_weights}. Proposition~\ref{prop:forward_weights_expectation} reveals that, in expectation, the forward weights are chosen so that their conditional expectation vanishes when $R^*=1$, i.e., truly positive cases do not require correction. Consequently, the explicit contribution to the correction term arises from samples with $R^*=0$. This structure allows us to recover the correct gradient direction, as shown in the following theorem.

\begin{theorem}[REINFORCE-Style Directional Alignment of the Forward Correction]
  \label{thm:forward_rho1_direction}
  Let the gradient-update term be $\Delta\theta = \frac{1}{M}\sum w_{\tilde R}\,G_t$, where $w_0=\rho_1-1$ and $w_1=\rho_1$ with $M$ trajectories. Under the Verifier Reward Channel model, the expected update is parallel to the clean policy gradient:
  \begin{equation}
    \mathbb{E}[\Delta\theta] \;=\; c\,\nabla_\theta J(\theta),
  \end{equation}
  in which $c=(1-\rho_0-\rho_1)$.
\end{theorem}
The proof is given in Appendix~\ref{app:proof_thm_forward_rho1}. Since $1-\rho_0-\rho_1 > 0$, Theorem~\ref{thm:forward_rho1_direction} guarantees that the expected update of the policy is in the same direction as the true gradient; meanwhile, the positive scaling factor $(1-\rho_0-\rho_1)$ can be absorbed into the learning rate. By avoiding the inverse operation of PGBC, PGFC circumvents the variance-inflation problem. 

\begin{algorithm}[t]
  \caption{Noisy Policy Gradient with Forward Correction (PGFC)}
  \label{alg:forward_rho1}
  \begin{algorithmic}
    \STATE \textbf{Input:} Initial policy $\theta_0$; learning rate $\eta$; batch size $M$; an estimate of the false negative rate, $\hat\rho_1$.
    \LOOP
    \STATE Roll out $M$ trajectories and observe rewards $\{\tilde R_i\}_{i=1}^M$.
    \STATE For each trajectory $i$, define the weight based on the observed reward $\tilde R_i$:
    \[
      w_{\tilde R_i} \;\leftarrow\;
      \begin{cases}
        \hat\rho_1-1, & \text{if } \tilde R_i=0,\\
        \hat\rho_1, & \text{if } \tilde R_i=1.
      \end{cases}
    \]
    \STATE For each trajectory $i$, compute the score-function term $G_i$ and the weighted term $h_i \leftarrow w_{\tilde R_i}\,G_i$.
    \STATE Form the policy gradient estimate: $\hat{g} \leftarrow \frac{1}{M}\sum_{i=1}^M h_i$.
    \STATE Update parameters: $\theta\leftarrow\theta + \eta\,\hat{g}$.
    \ENDLOOP
  \end{algorithmic}
\end{algorithm}

\begin{figure*}[t!]
  \captionsetup[subfigure]{font={color=black}}
  \centering
  \subfloat[Qwen2.5-Math-1.5B]{
    \includegraphics[width=0.48\linewidth]{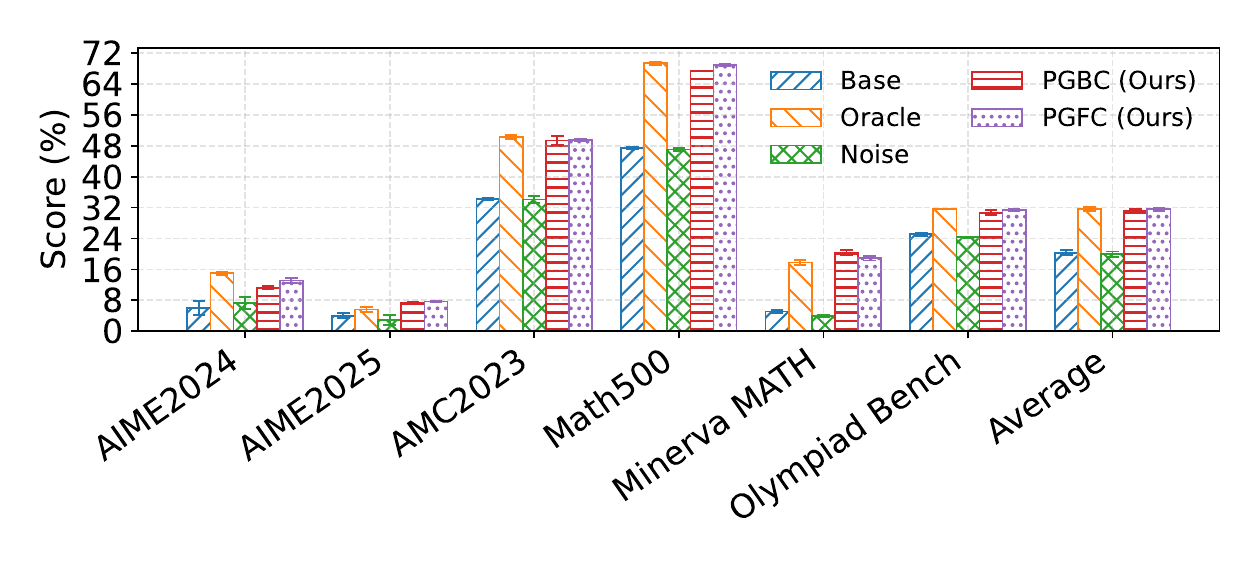}
  }\hfill
  \subfloat[DeepSeek-R1-Distill-Qwen-1.5B]{
    \includegraphics[width=0.48\linewidth]{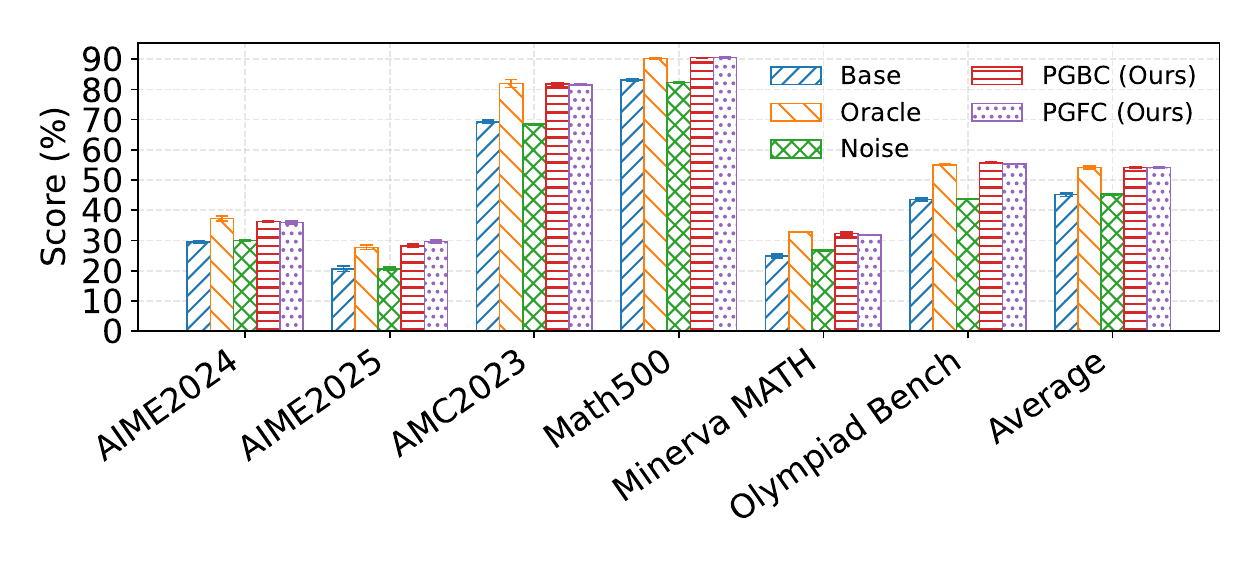}
  }\vfill
  \subfloat[Llama-3.2-3B-Instruct]{
    \includegraphics[width=0.48\linewidth]{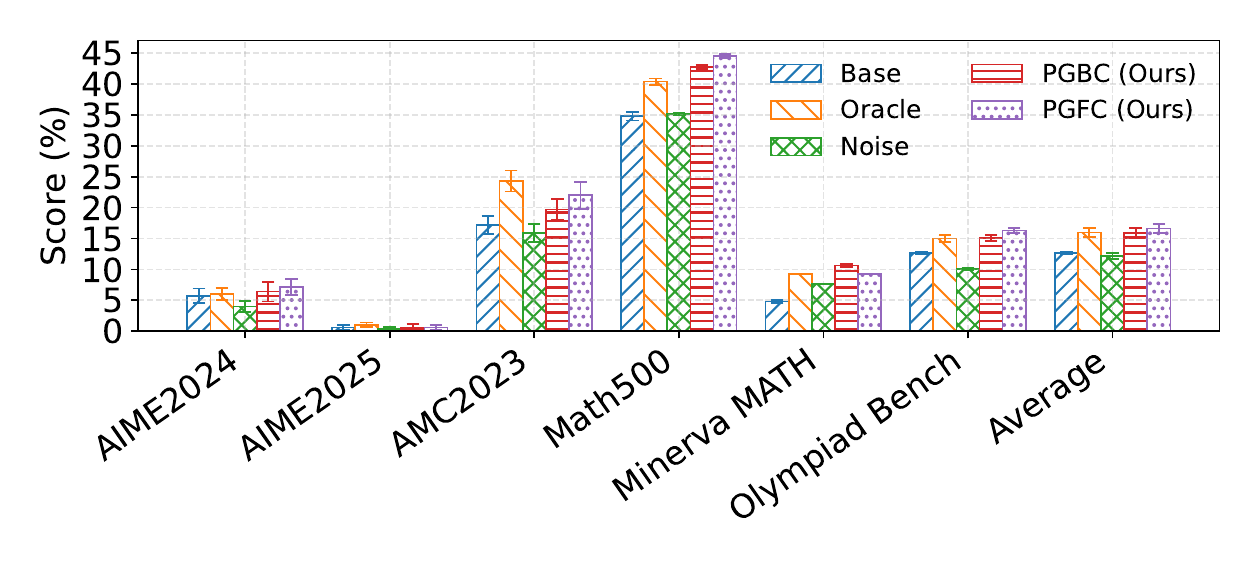}
  }\hfill
  \subfloat[Qwen2.5-Math-7B]{
    \includegraphics[width=0.48\linewidth]{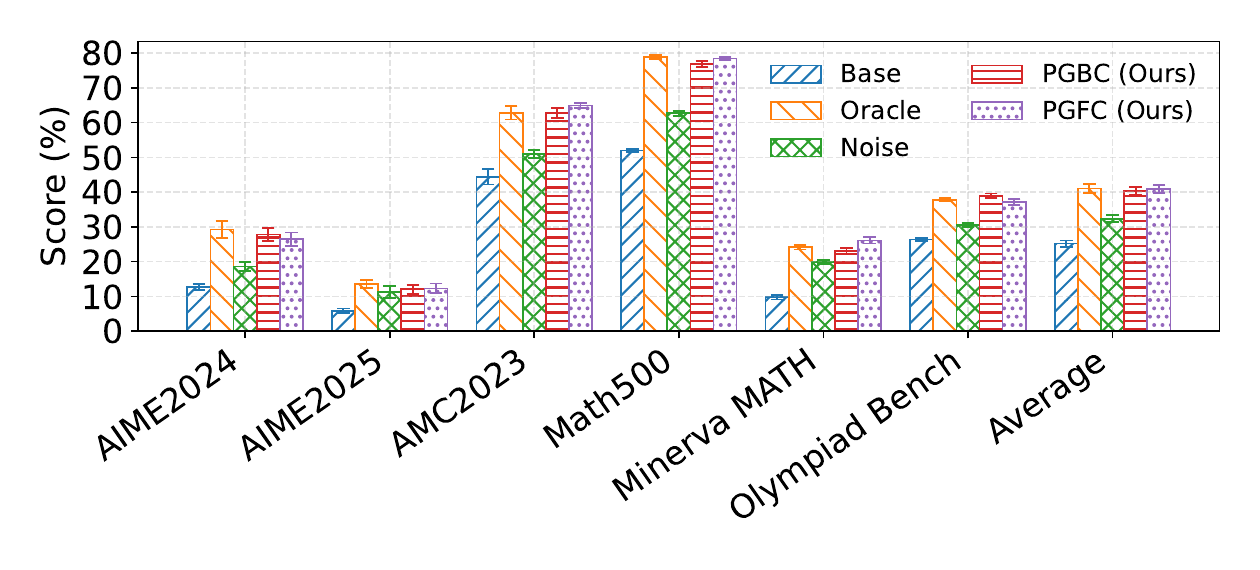}
  }
  \caption{Synthetic-Noise Results \textbf{(pass@1)} with 16 samples and 5 random seeds. \textcolor{blue}{\textbf{Base}}: baseline without RL; \textcolor{orange}{\textbf{Oracle}}: Training with clean rewards; \textcolor{green!60!black}{\textbf{Noise}}: Training with noisy verifier rewards; \textcolor{red}{\textbf{Noise\_BC}}: Training with noise under backward correction; \textcolor{violet}{\textbf{Noise\_FC}}: Training with noise under forward correction.}
  \label{fig:synthetic-noise-bars}
\end{figure*}

In practice, we estimate \(\hat\rho_1\) during training by \emph{appealing} a small, uniformly random subset of rule-based negatives to a lightweight LLM verifier (e.g., TinyV~\citep{xu2025tinyv}) and then smoothing the empirical flip rate. Concretely, at step \(t\), let \(\mathcal N_{\mathrm R}^{(t)}\) be items labeled negative by the rule-based verifier, \(\mathcal P_{\mathrm R}^{(t)}\) the rule-based positives, and sample a fraction \(q\in(0,1]\) of \(\mathcal N_{\mathrm R}^{(t)}\) for appeal; denote by \(\mathcal P_{\mathrm L}^{(t)}\) those appealed items that the LLM flips to positive. Using a Horvitz--Thompson correction~\citep{karwa2023admissibility}, we estimate the number of FNs as \(|\mathcal P_{\mathrm L}^{(t)}|/q\), treat \(|\mathcal P_{\mathrm R}^{(t)}|\) as true positives in the rule-checker regime, and set \(\hat\rho_1^{(t)}=\frac{|\mathcal P_{\mathrm L}^{(t)}|/q+\alpha}{|\mathcal P_{\mathrm L}^{(t)}|/q+|\mathcal P_{\mathrm R}^{(t)}|+\alpha+\beta}\) with small Beta-prior pseudocounts \(\alpha,\beta=10^{-5}\). We then apply exponential moving average (EMA) smoothing over a sliding window. If both \(\hat\rho_0\) and \(\hat\rho_1\) are available, one can use PGBC; if only \(\hat\rho_1\) is available, PGFC is the default. We released our code in ~\url{https://github.com/caixq1996/noisy-RLVR}. More details of appealing mechanism can be found in Appendix~\ref{app:rho1-hparams-ablation}.

\paragraph{Bridging REINFORCE-Style Corrections to GRPO Updates}

Our analysis is stated for the reward-gradient component of a score-function estimator, while GRPO implements this component through scalar trajectory coefficients broadcast as token-level advantages inside a clipped surrogate. We therefore insert each correction at the coefficient level that best matches its estimator: PGBC is implemented as a corrected-utility coefficient based on \(\widehat r_i=(\tilde r_i-\hat\rho_0)/(1-\hat\rho_0-\hat\rho_1)\), whereas PGFC is implemented as a score-level coefficient \(w_{\tilde r_i}\in\{\hat\rho_1-1,\hat\rho_1\}\), not as a reward proxy. In both cases, the corrected coefficient is passed directly to the GRPO surrogate, with the KL term unchanged. This implementation avoids a subtle binary-reward degeneracy: any pre-normalization proxy affine in \(\tilde r_i\) can be removed by group centering and standardization, making the corrected update identical to uncorrected noisy-reward GRPO. Appendix~\ref{app:REINFORCE-to-GRPO} states the centered-update guarantee and Appendix~\ref{app:impl-grpo} gives the implementation recipe.

\section{Experiments}
\label{sec:experiments}
We evaluate our approach under both \emph{synthetic} and \emph{real-world} verifier noise. We first spell out the experimental protocol, and then present main and ablation results. Additional experimental results, including training dynamics, pass@8 performance under real-world noise, appeals statistics, complementarity with stronger verifiers, and non-math evaluations are provided in Appendix~\ref{app:additional-experiments}.


\subsection{Experimental Setup}
\label{sec:exp-setup}
We train on three small backbones, \textit{Qwen2.5-Math-1.5B}, \textit{DeepSeek-R1-Distill-Qwen-1.5B}, and \textit{Llama-3.2-3B-Instruct}, and probe scale with \textit{Qwen2.5-Math-7B}. Rewards come from either (i) a rule-based checker that extracts the final \verb|\boxed{\cdot}| answer and tests numeric/rational equivalence or (ii) a lightweight LLM verifier (\textbf{TinyV 1.5B}) used for appeals/estimation; prior work motivates explicit noise modeling due to systematic FP/FN behavior in these verifiers. Unless stated otherwise, we follow the \texttt{DeepScaleR} corpus and a GRPO-style on-policy recipe with BoN sampling, implementing our \emph{backward} and \emph{forward} corrections as drop-in hooks at the advantage-construction stage within \texttt{VERL}. Evaluation uses six verifiable math suites—\textsc{AIME-2024}, \textsc{AIME-2025}, \textsc{AMC-2023}, \textsc{MATH500}, \textsc{Minerva MATH}, and \textsc{OlympiadBench}—reporting \emph{Pass@1} and \emph{Pass@8} with 16 samples and 5 random seeds in average. Compute is 8$\times$A100 (40GB) GPUs servers; unless noted, KL schedules, sampling temperatures, and other rollout settings are kept identical across compared conditions. 
We estimate pass@$k$ as $1-\binom{n-c_{\rm ok}}{k}/\binom{n}{k}$, where $n=16$, $c_{\rm ok}$ is the number of correct samples, and $k\in\{1,8\}$~\citep{chen2021evaluatinglargelanguagemodels}. More details can be found in Appendix~\ref{app:prompts-and-hparams}.

\subsection{Synthetic Noise}
\label{subsec:synthetic-noise-exp}
\begin{figure*}[t!]
  \captionsetup[subfigure]{font={color=black}}
  \color{black}\captionsetup{font={color=black}}\centering
  \subfloat[Qwen2.5-Math-1.5B]{
    \includegraphics[width=0.48\linewidth]{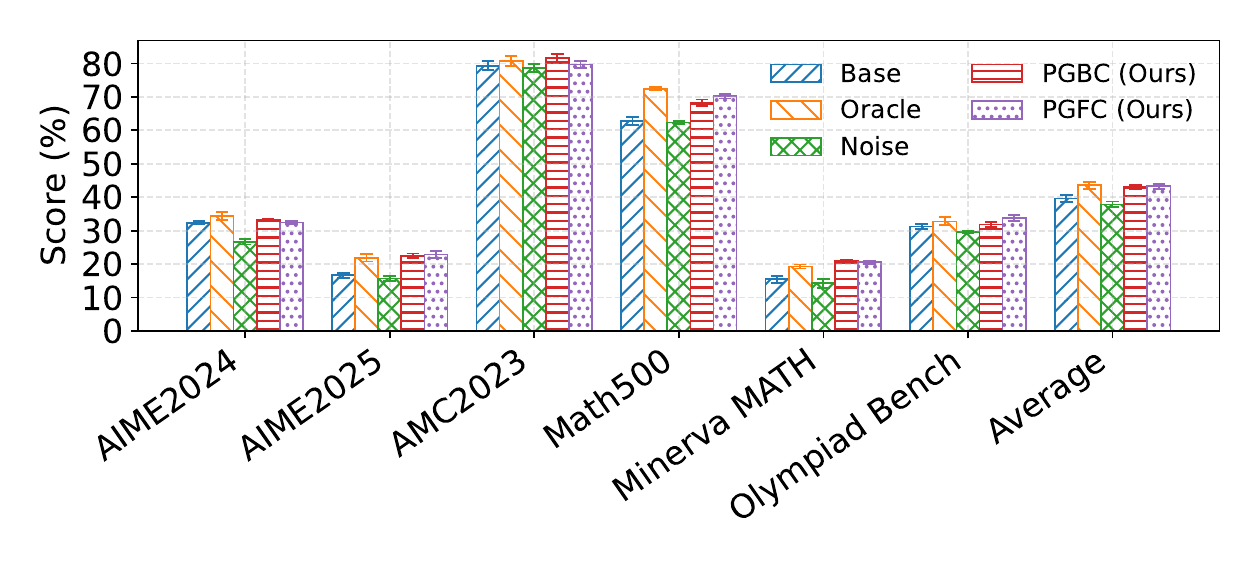}
  }\hfill
  \subfloat[DeepSeek-R1-Distill-Qwen-1.5B]{
    \includegraphics[width=0.48\linewidth]{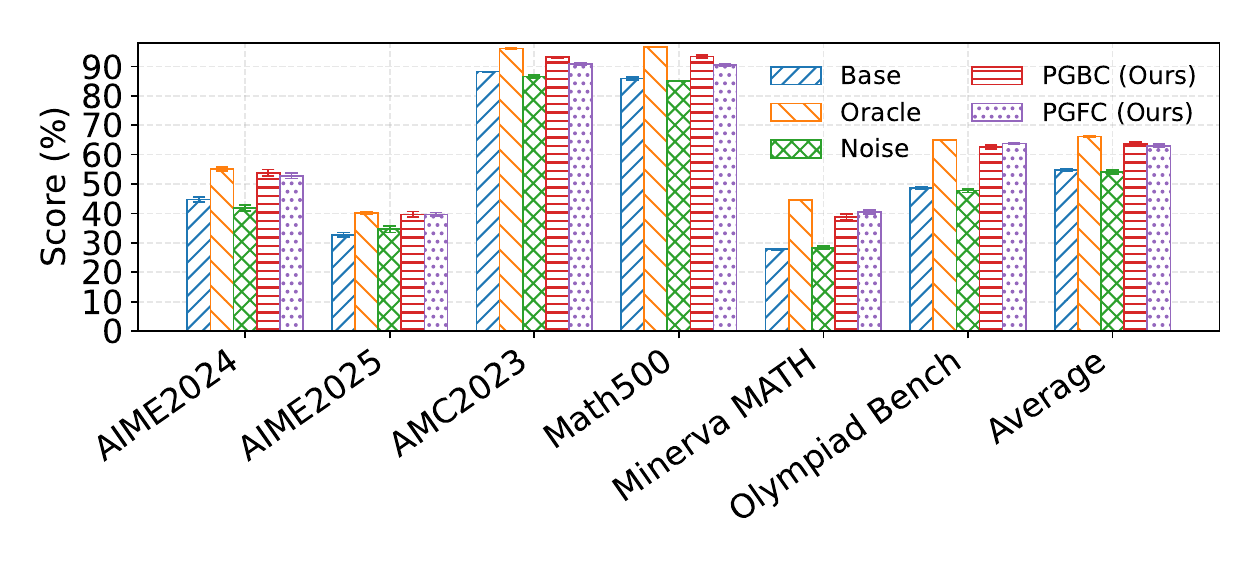}
  }\vfill
  \subfloat[Llama-3.2-3B-Instruct]{
    \includegraphics[width=0.48\linewidth]{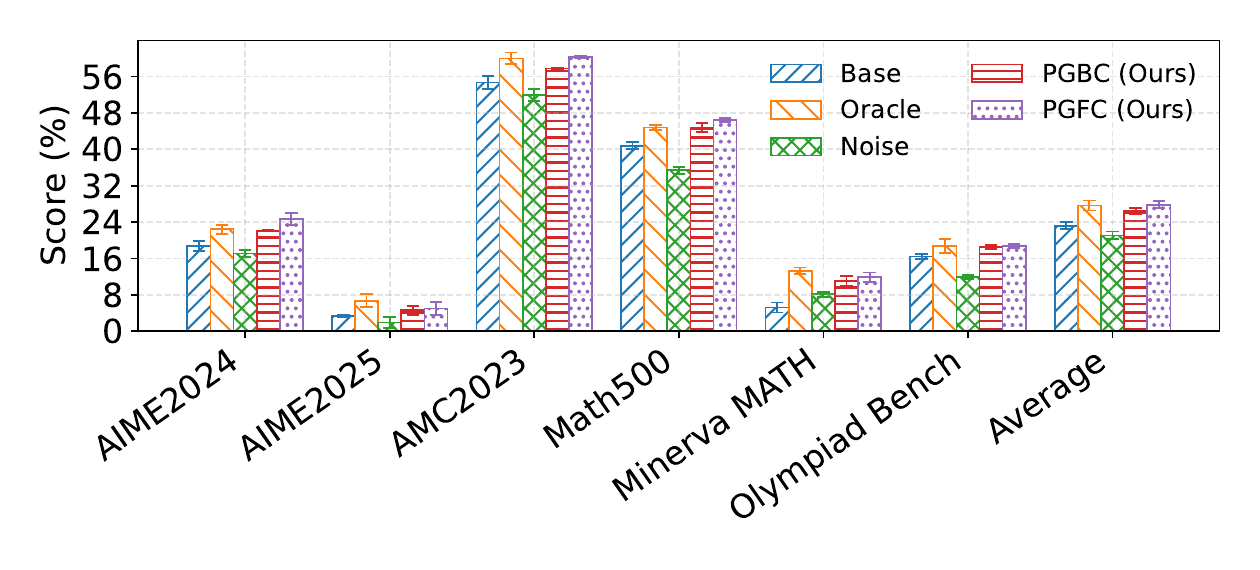}
  }\hfill
  \subfloat[Qwen2.5-Math-7B]{
    \includegraphics[width=0.48\linewidth]{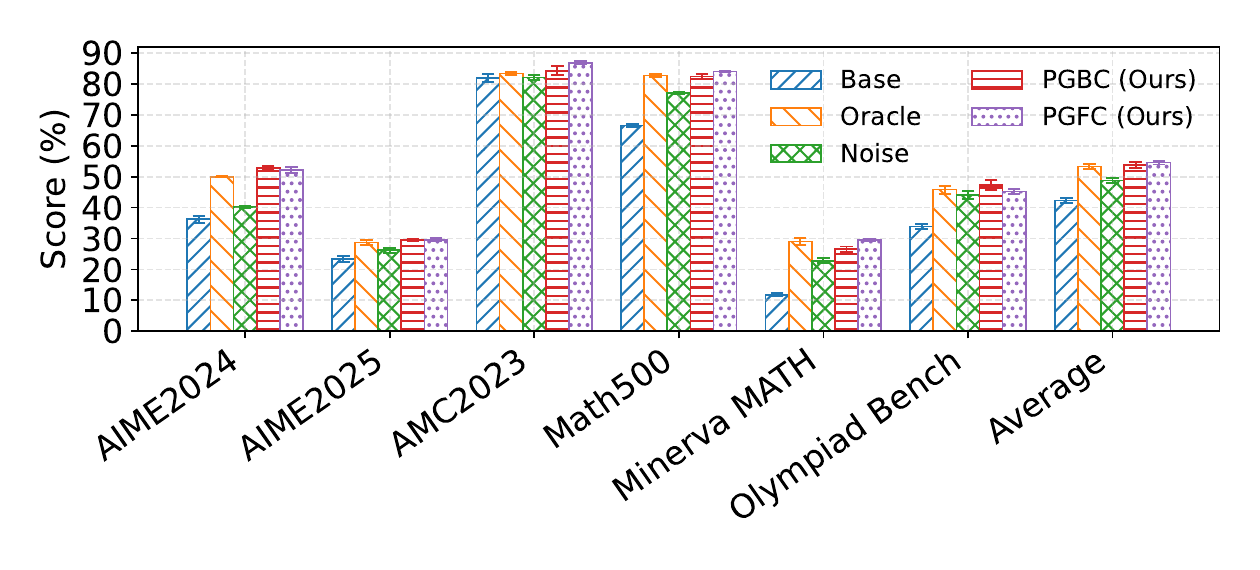}
  }
  \caption{Synthetic-Noise Results \textbf{(pass@8)} with 16 samples and 5 random seeds. \textcolor{blue}{\textbf{Base}}: baseline without RL; \textcolor{orange}{\textbf{Oracle}}: Training with clean rewards; \textcolor{green!60!black}{\textbf{Noise}}: Training with noisy verifier rewards; \textcolor{red}{\textbf{Noise\_BC}}: Training with noise under backward correction; \textcolor{violet}{\textbf{Noise\_FC}}: Training with noise under forward correction.}
  \vspace{-4mm}
  \label{fig:synthetic-noise-bars-pass8}
\end{figure*}

To disentangle optimization effects from verifier unreliability, we inject \emph{verifier-conditional} noise into the binary reward stream during training. Concretely, when the latent clean reward is $R^*\!\in\{0,1\}$, the observed reward $\tilde R$ is drawn from a reward channel with $\Pr(\tilde R{=}1\!\mid\!R^*{=}0)=\rho_0$ and $\Pr(\tilde R{=}0\!\mid\!R^*{=}1)=\rho_1$; unless stated otherwise we use $\rho_0{=}0.1$ and $\rho_1{=}0.2$. We train GRPO on DeepScaleR with identical sampling and KL settings across conditions, and evaluate Pass@1 (16 samples) and Pass@8 on \textsc{AIME-2024/2025}, \textsc{AMC-2023}, \textsc{MATH500}, \textsc{Minerva MATH}, and \textsc{OlympiadBench}. We compare five variants: \emph{Base} (no RL), \emph{Oracle} (clean rewards), \emph{Noise} (uncorrected), and our two corrections—\emph{PGBC} (backward correction using $\widehat R=\frac{\tilde R-\rho_0}{1-\rho_0-\rho_1}$) and \emph{PGFC} (forward correction that rescales the gradient using $\rho_1$ only).
Figure~\ref{fig:synthetic-noise-bars-pass8} complements the pass@1 results with synthetic-noise pass@8: on the Qwen and Llama backbones, PGBC and PGFC improve over Noise and remain close to Oracle in the six-benchmark average.

\subsection{Real-World Noise}
\label{subsec:real-world-exp}
As discussed above, automated verifiers exhibit both \emph{false positives} (LLM judges over-crediting incorrect solutions) and \emph{false negatives} (rule-based checkers rejecting correct ones) \citep{LLMJudgeSurvey2024,xu2025tinyv,huang2025pitfalls,Li2025VerifyBench}. In math RL pipelines, the latter is particularly prevalent: brittle exact-match or limited equivalence rules lead to many valid answers being scored as incorrect, depriving the agent of learning signal. Motivated by this, in this subsection we specifically investigate whether denoising \emph{false negatives}, i.e., estimating $\hat\rho_1$ as in Section~\ref{subsec:algo_forward}, can measurably improve policy optimization. Concretely, we keep a fast rule-based checker as the \emph{primary} reward source with a lightweight LLM verifier (TinyV) \citep{xu2025tinyv}. From disagreements on negatives we maintain an online estimate $\hat\rho_1$ (EMA over a sliding window to track policy drift).

\vspace{-2mm}
Results in Table~\ref{tab:realworld-all} show a consistent pattern across backbones and benchmarks. Direct LLM-judge rewards underperform Rule on the six-benchmark average for each backbone, although individual task outcomes vary. Employing appeals to recheck rule-based negatives (\texttt{Appeals}) recovers some FNs, but its downstream benefit varies across models and benchmarks. In contrast, \texttt{Appeals+PGFC} uses the same appeals stream to estimate \(\hat\rho_1\) and directly applies the FN correction to the policy gradient via weights \(w_{\tilde R}\!\in\!\{\hat\rho_1-1,\hat\rho_1\}\), improving over Appeals in 22 of 24 cells.

\begin{table*}[t]
  \centering
  \small
  \caption{Real-world pass@1. \texttt{Rule}: uncorrected rule-based rewards; \texttt{LLM-as-Judge}: direct LLM-judge rewards; \texttt{Appeals}: rule-based reward plus LLM appeals on negative samples without gradient correction; \texttt{Appeals+PGFC}: appeals-based FN-rate estimation plus forward correction.}
  \label{tab:realworld-all}\label{tab:deepseek-official-additional}
  \resizebox{\linewidth}{!}{
    \begin{tabular}{lrrrrrr|r}
      \toprule
      Dataset & AIME2024 & AIME2025 & AMC2023 & MATH500 & Minerva MATH & Olympiad Bench & Average \\
      \midrule
      \multicolumn{8}{c}{Qwen2.5-Math-1.5B} \\
      Base & 6.0 $\pm$ 1.9 & 4.0 $\pm$ 0.6 & 34.2 $\pm$ 0.2 & 47.5 $\pm$ 0.3 & 5.1 $\pm$ 0.4 & 25.1 $\pm$ 0.4 & 20.3 $\pm$ 0.6 \\
      Rule & 15.0 $\pm$ 0.4 & 5.6 $\pm$ 0.6 & 50.3 $\pm$ 0.6 & \textbf{69.4 $\pm$ 0.4} & \textbf{17.8 $\pm$ 0.6} & 31.6 $\pm$ 0.0 & 31.6 $\pm$ 0.4 \\
      LLM-as-Judge & 10.9 $\pm$ 1.3 & 4.7 $\pm$ 1.0 & 42.1 $\pm$ 1.8 & 63.0 $\pm$ 0.7 & 15.9 $\pm$ 0.7 & 25.3 $\pm$ 0.5 & 27.0 $\pm$ 1.0 \\
      Appeals & 11.9 $\pm$ 0.6 & 5.8 $\pm$ 1.2 & 47.8 $\pm$ 1.2 & 68.3 $\pm$ 0.1 & 16.7 $\pm$ 0.6 & 29.8 $\pm$ 0.1 & 30.1 $\pm$ 0.6 \\
      Appeals+PGFC (Ours) & \textbf{20.3 $\pm$ 0.0} & \textbf{10.7 $\pm$ 1.7} & \textbf{53.3 $\pm$ 1.4} & 68.6 $\pm$ 0.8 & 16.5 $\pm$ 0.4 & \textbf{32.9 $\pm$ 0.2} & \textbf{33.7 $\pm$ 0.8} \\
      \midrule
      \multicolumn{8}{c}{\rev{DeepSeek-R1-Distill-Qwen-1.5B}} \\
      \rev{Base} & \rev{29.5 $\pm$ 0.4} & \rev{20.6 $\pm$ 0.9} & \rev{69.3 $\pm$ 0.6} & \rev{83.1 $\pm$ 0.3} & \rev{24.9 $\pm$ 0.6} & \rev{43.6 $\pm$ 0.5} & \rev{45.2 $\pm$ 0.6} \\
      \rev{Rule} & \rev{31.0 $\pm$ 0.7} & \rev{26.3 $\pm$ 0.5} & \rev{79.1 $\pm$ 0.9} & \rev{88.7 $\pm$ 0.4} & \rev{29.5 $\pm$ 0.1} & \rev{50.1 $\pm$ 0.2} & \rev{50.8 $\pm$ 0.5} \\
      \rev{LLM-as-Judge} & \rev{30.2 $\pm$ 1.0} & \rev{27.0 $\pm$ 0.8} & \rev{79.5 $\pm$ 0.9} & \rev{85.8 $\pm$ 0.7} & \rev{27.7 $\pm$ 0.9} & \rev{47.2 $\pm$ 1.0} & \rev{49.6 $\pm$ 0.9} \\
      \rev{Appeals} & \rev{34.0 $\pm$ 1.2} & \rev{27.8 $\pm$ 0.9} & \rev{78.7 $\pm$ 1.1} & \rev{88.0 $\pm$ 0.8} & \rev{29.0 $\pm$ 1.0} & \rev{49.6 $\pm$ 1.1} & \rev{51.2 $\pm$ 1.0} \\
      \rev{Appeals+PGFC (Ours)} & \rev{\textbf{35.8 $\pm$ 0.7}} & \rev{\textbf{29.1 $\pm$ 0.8}} & \rev{\textbf{81.7 $\pm$ 0.6}} & \rev{\textbf{88.8 $\pm$ 0.5}} & \rev{\textbf{31.1 $\pm$ 0.7}} & \rev{\textbf{52.4 $\pm$ 0.6}} & \rev{\textbf{53.2 $\pm$ 0.7}} \\
      \midrule
      \multicolumn{8}{c}{Llama-3.2-3B-Instruct} \\
      Base & 5.7 $\pm$ 1.2 & 0.6 $\pm$ 0.4 & 17.2 $\pm$ 1.5 & 34.8 $\pm$ 0.7 & 4.8 $\pm$ 0.3 & 12.7 $\pm$ 0.2 & 12.6 $\pm$ 0.2 \\
      Rule & 6.0 $\pm$ 1.0 & \textbf{1.0 $\pm$ 0.4} & \textbf{24.3 $\pm$ 1.7} & 40.4 $\pm$ 0.5 & 8.9 $\pm$ 0.7 & 15.0 $\pm$ 0.5 & 15.9 $\pm$ 0.8 \\
      LLM-as-Judge & 1.4 $\pm$ 1.1 & 0.1 $\pm$ 0.3 & 12.7 $\pm$ 2.1 & 26.4 $\pm$ 0.2 & 4.1 $\pm$ 0.2 & 7.7 $\pm$ 0.4 & 8.7 $\pm$ 0.6 \\
      Appeals & 3.8 $\pm$ 1.4 & 0.6 $\pm$ 0.7 & 16.9 $\pm$ 1.8 & 40.8 $\pm$ 0.3 & 9.0 $\pm$ 0.4 & 13.7 $\pm$ 0.3 & 14.1 $\pm$ 0.4 \\
      Appeals+PGFC (Ours) & \textbf{9.4 $\pm$ 1.9} & 0.4 $\pm$ 0.6 & 21.2 $\pm$ 1.3 & \textbf{46.4 $\pm$ 0.3} & \textbf{9.9 $\pm$ 0.8} & \textbf{16.3 $\pm$ 0.3} & \textbf{17.3 $\pm$ 0.9} \\
      \midrule
      \multicolumn{8}{c}{Qwen2.5-Math-7B} \\
      Base & 12.7 $\pm$ 0.9 & 5.8 $\pm$ 0.7 & 44.4 $\pm$ 2.2 & 52.0 $\pm$ 0.4 & 9.8 $\pm$ 0.7 & 26.4 $\pm$ 0.4 & 25.2 $\pm$ 0.9 \\
      Rule & 29.2 $\pm$ 2.5 & 13.5 $\pm$ 1.2 & 62.8 $\pm$ 1.9 & 78.9 $\pm$ 0.5 & 24.2 $\pm$ 0.7 & 37.8 $\pm$ 0.5 & 41.1 $\pm$ 1.2 \\
      LLM-as-Judge & 16.8 $\pm$ 2.1 & 6.8 $\pm$ 1.4 & 50.2 $\pm$ 1.8 & 62.5 $\pm$ 1.0 & 10.1 $\pm$ 0.4 & 31.0 $\pm$ 0.9 & 29.6 $\pm$ 1.3 \\
      Appeals & 27.4 $\pm$ 1.9 & 11.8 $\pm$ 1.9 & 63.7 $\pm$ 1.6 & 74.9 $\pm$ 0.7 & 20.6 $\pm$ 0.8 & 37.6 $\pm$ 0.3 & 39.3 $\pm$ 1.2 \\
      Appeals+PGFC (Ours) & \textbf{31.0 $\pm$ 1.5} & \textbf{14.6 $\pm$ 1.2} & \textbf{65.7 $\pm$ 1.4} & \textbf{81.6 $\pm$ 0.4} & \textbf{26.2 $\pm$ 0.5} & \textbf{39.3 $\pm$ 0.5} & \textbf{43.1 $\pm$ 0.9} \\
      \bottomrule
    \end{tabular}
  }
\end{table*}

\subsection{Robustness to Noise Mis-Specification}
\label{subsec:robustness}

\begin{figure*}[t]
  \centering

  \subfloat[BC: $\hat{\rho}_0{=}0.1$, $\hat{\rho}_1\!\in\!\lbrack 0.1,0.3\rbrack$]{%
    \begin{minipage}[b]{0.31\linewidth}\centering
      \includegraphics[width=1.\linewidth]{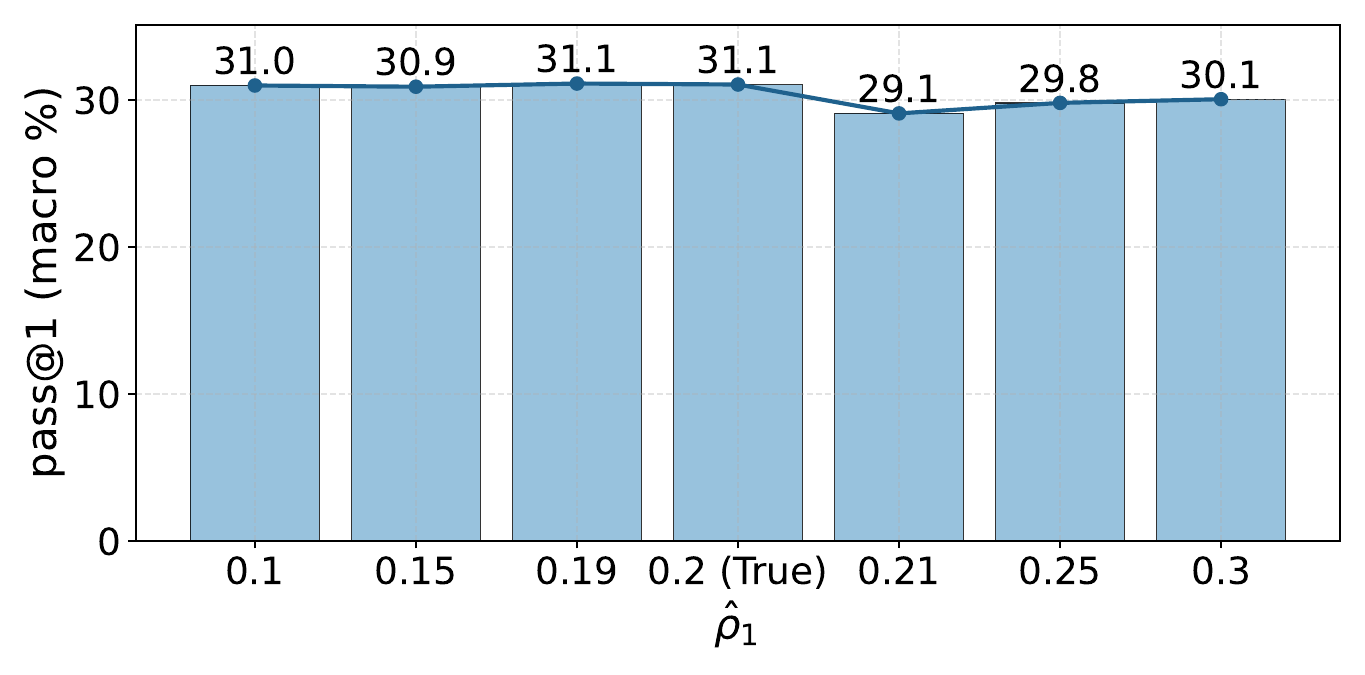}
    \end{minipage}
  }
  \hfill
  \subfloat[BC: $\hat{\rho}_1{=}0.2$, $\hat{\rho}_0\!\in\!\lbrack 0.0,0.2\rbrack$]{%
    \begin{minipage}[b]{0.31\linewidth}\centering
      \includegraphics[width=1.\linewidth]{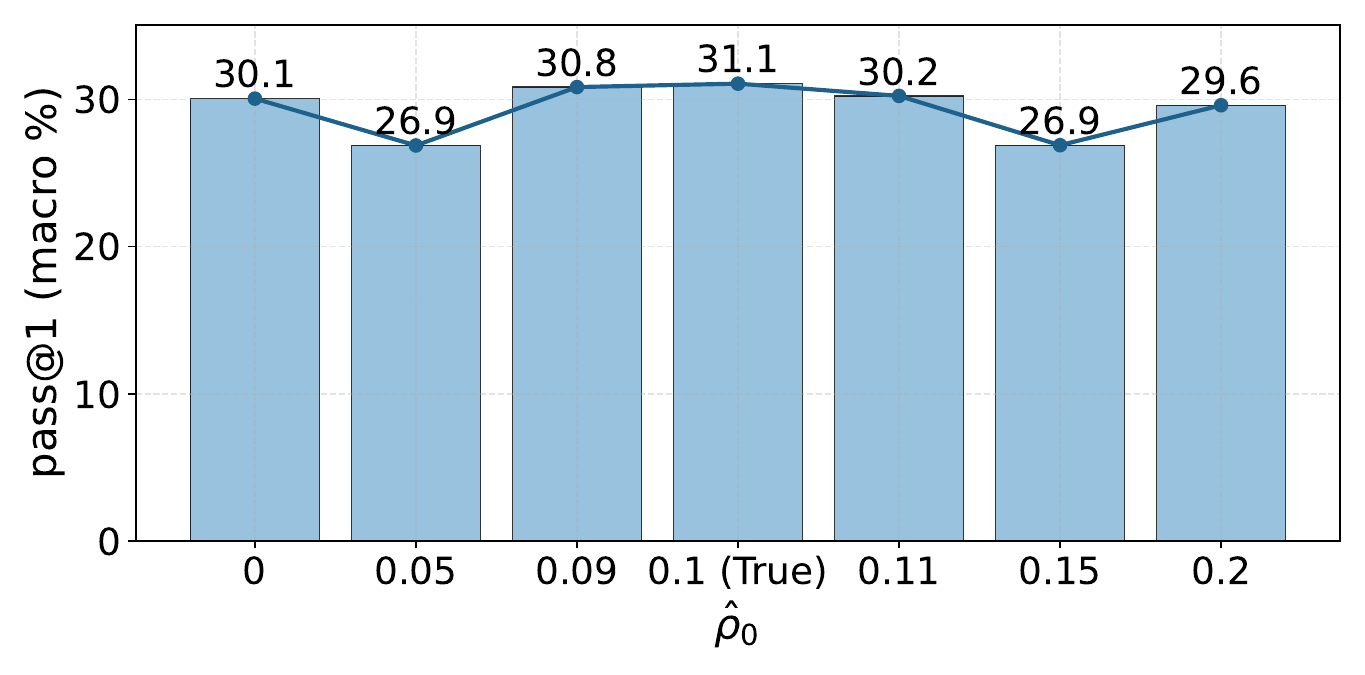}
    \end{minipage}
  }
  \hfill
  \subfloat[FC: $\hat{\rho}_0{=}0.1$, $\hat{\rho}_1\!\in\!\lbrack 0.1,0.3\rbrack$]{%
    \begin{minipage}[b]{0.31\linewidth}\centering
      \includegraphics[width=1.\linewidth]{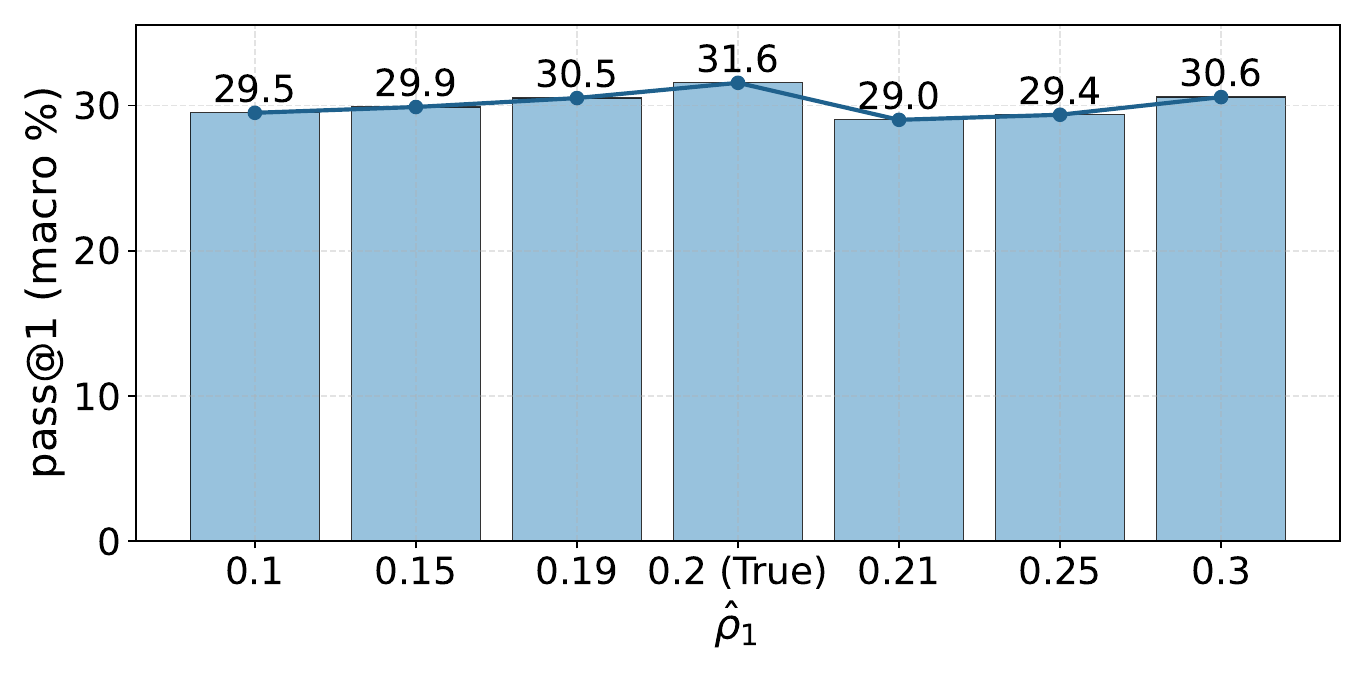}
    \end{minipage}
  }
  \caption{Robustness results. (a) Backward correction (BC) with $\hat{\rho}_0$ fixed and sweeping $\hat{\rho}_1$; (b) Backward correction (BC) with $\hat{\rho}_1$ fixed and sweeping $\hat{\rho}_0$; (c) Forward correction (FC) with $\hat{\rho}_0$ fixed and sweeping $\hat{\rho}_1$.}
  \label{fig:robustness}
\end{figure*}

There now exist practical procedures to \emph{estimate} verifier FP/FN rates via rule-based equivalence checkers (e.g., \textsc{Math-Verify}), lightweight appeals/judges such as TinyV, and meta-evaluation suites that quantify verifier reliability \citep{math-verify,xu2025tinyv,Li2025VerifyBench}. In real deployments, however, these estimates can be imperfect. We therefore test how \emph{backward} and \emph{forward} corrections behave under mis-specified noise rates. We follow the synthetic-noise setting from Section~\ref{subsec:synthetic-noise-exp} with Qwen2.5-math-1.5B: rewards are \emph{corrupted at data-time} with \(\rho_0{=}0.1,\ \rho_1{=}0.2\). During training, we \emph{intentionally} feed each algorithm mis-specified rates from a grid \(\hat\rho_0 \in [0.0, 0.2]\), \(\hat\rho_1 \in [0.1, 0.3]\), and report the Average Pass@1 (16 samples) across our six math benchmarks. Because the forward method only requires the false negative rate, we vary \(\hat\rho_1\) for forward correction while keeping \(\hat\rho_0\) unused, as shown in Fig.~\ref{fig:robustness}.

\textit{Backward} correction remains strong when the total noise is underestimated, but performance degrades as we overestimate the rates (i.e., as $1-\hat\rho_0-\hat\rho_1$ shrinks). Intuitively, the unbiased de-biasing,
\(\widehat R = \frac{\tilde R - \hat\rho_0}{1-\hat\rho_0-\hat\rho_1}\),
involves an inverse factor that amplifies reward variance and accentuates estimation error when the effective denominator becomes small. In contrast, \textit{forward} correction exhibits a notably flat response curve around the true $\rho_1$; perturbing $\hat\rho_1$ changes only the per-sample weights and avoids any inverse operation, yielding lower-variance policy gradient estimates.

We also stress-test the constant-channel assumption using format-dependent, \textbf{non-iid verifier noise} with the same Qwen2.5-Math-1.5B / DeepScaleR setup. Both the FN and FP rates depend on a prompt-level tag and a response-level format tag (e.g., decimal or missing boxed answer vs.\ canonical format). Specifically, we use a shared FP rate of \(\rho_0=0.2\) but a format-dependent FN rate: \(\rho_1=0.05\) for low-risk formats and \(\rho_1=0.45\) for high-risk formats, with the actual flip decision made by hashing \texttt{(data\_source, ground\_truth, solution\_str)}. This explicitly violates the instance-independent channel model. Table~\ref{tab:format-dependent-noise} shows that both corrections still improve over uncorrected training and base model under this non-iid noise setting, with PGFC again strongest performance in most cases.

\begin{table*}[t]
  \centering
  \small
  \caption{Format-dependent \textbf{non-iid verifier-noise} setting on Qwen2.5-Math-1.5B. Each cell reports pass@1 / pass@8.}
  \label{tab:format-dependent-noise}
  \resizebox{\linewidth}{!}{
    \begin{tabular}{lccccccc}
      \toprule
      Method & MATH500 & AIME2024 & AIME2025 & AMC2023 & Minerva Math & OlympiadBench & Average \\
      \midrule
      Base & 47.5 / 62.8 & 6.0 / 32.4 & 4.0 / 16.7 & 34.2 / 79.3 & 5.1 / 15.5 & 25.1 / 31.3 & 20.3 / 39.7 \\
      Noise & 40.7 / 60.5 & 7.5 / 26.5 & 2.2 / 15.8 & 30.0 / 72.4 & 3.8 / 11.6 & 21.5 / 25.8 & 17.6 / 35.4 \\
      PGBC & 67.7 / 67.9 & 11.5 / \textbf{33.0} & 7.4 / 22.4 & 49.1 / \textbf{81.8} & 20.7 / 21.1 & 30.5 / 31.7 & 31.2 / 43.0 \\
      PGFC & \textbf{68.1} / \textbf{69.9} & \textbf{12.8} / 32.4 & \textbf{8.0} / \textbf{22.6} & \textbf{49.6} / 79.5 & \textbf{21.8} / \textbf{23.8} & \textbf{31.1} / \textbf{33.6} & \textbf{31.9} / \textbf{43.6} \\
      \bottomrule
    \end{tabular}
  }
  \vspace{-4mm}
\end{table*}

\section{Conclusion}
\label{sec:conclusion}
\vspace{-2mm}
Reinforcement Learning with Verifiable Rewards (RLVR) has emerged as a practical and popular paradigm for training reasoning-focused large language models (LLMs). Yet binarized rewards inevitably suffer from \emph{false positives} (FP) and \emph{false negatives} (FN). Prior efforts largely mitigated these errors from the verifier side—e.g., debiasing LLM judges or strengthening equivalence checking—to reduce the downstream impact on learning \citep{Li2025VerifyBench,huang2025pitfalls}. In contrast, we modeled verifier unreliability as a verifier-conditional noise channel with the FP and FN rates, and introduced two theory-driven corrections that act directly at the policy-learning interface. The \textit{backward} correction de-biases the observed reward to yield an unbiased gradient estimator; it requires estimates of the FP and FN rates but is agnostic to the choice of policy-optimization algorithm and can thus be used beyond policy gradient methods. The \textit{forward} correction rescales score-function terms so that the expected update is aligned with the clean gradient; notably, it needs only the FN rate estimate, which is often the dominant source of real-world noise due to rule-based FNs \citep{huang2025pitfalls,xu2025tinyv}.

While we instantiated appeals with a lightweight LLM verifier for efficiency, the same backward/forward correction mechanisms should also gain from stronger verifiers~\citep{xu2025tinyv}. Meanwhile, although modeled with the constant noise rates, the proposed methods already show clear advantages, even under a more challenging non-iid verifier noise setting. On the other hand, real-world noise can depend on content, formatting, prompt style, and verifier family, motivating future work on instance-dependent reliability models and adversarially robust verifiers.

\bibliographystyle{iclr2026_conference}

\bibliography{arxiv_v4_references}

@article{RLVR2025,
  title   = {Reinforcement Learning with Verifiable Rewards Implicitly Incentivizes Correct Reasoning in Base LLMs},
  author  = {Xumeng Wen and
             Zihan Liu and
             Shun Zheng and
             Zhijian Xu and
             Shengyu Ye and
             Zhirong Wu and
             Xiao Liang and
             Yang Wang and
             Junjie Li and
             Ziming Miao and
             Jiang Bian and
             Mao Yang},
  journal = {arXiv preprint arXiv:2506.14245},
  year    = {2025}
}

@inproceedings{ICML:Li+etal:2021,
  author    = {Xuefeng Li and
               Tongliang Liu and
               Bo Han and
               Gang Niu and
               Masashi Sugiyama},
  editor    = {Marina Meila and
               Tong Zhang},
  title     = {Provably End-to-end Label-noise Learning without Anchor Points},
  booktitle = {Proceedings of the 38th International Conference on Machine Learning,
               {ICML} 2021, 18-24 July 2021, Virtual Event},
  series    = {Proceedings of Machine Learning Research},
  volume    = {139},
  pages     = {6403--6413},
  publisher = {{PMLR}},
  year      = {2021}
}

@article{xu2025tinyv,
  title   = {TinyV: Reducing False Negatives in Verification Improves RL for LLM Reasoning},
  author  = {Xu, Zhangchen and Li, Yuetai and Jiang, Fengqing and Ramasubramanian, Bhaskar and Niu, Luyao and Lin, Bill Yuchen and Poovendran, Radha},
  journal = {arXiv preprint arXiv:2505.14625},
  year    = {2025}
}

@misc{DeepScaleR,
  title        = {DeepScaleR: Surpassing O1-Preview with a 1.5B Model by Scaling RL},
  author       = {Michael Luo and Sijun Tan and Justin Wong and Xiaoxiang Shi and William Y. Tang and Manan Roongta and Colin Cai and Jeffrey Luo and Li Erran Li and Raluca Ada Popa and Ion Stoica},
  howpublished = {\url{https://pretty-radio-b75.notion.site/DeepScaleR-Surpassing-O1-Preview-with-a-1-5B-Model-by-Scaling-RL-19681902c1468005bed8ca303013a4e2}},
  note         = {Notion Blog},
  year         = {2025}
}

@article{karwa2023admissibility,
  title   = {On the admissibility of horvitz-thompson estimator for estimating causal effects under network interference},
  author  = {Karwa, Vishesh and Airoldi, Edoardo M},
  journal = {arXiv preprint arXiv:2312.01234},
  year    = {2023}
}

@inproceedings{minerva,
  author    = {Aitor Lewkowycz and
               Anders Andreassen and
               David Dohan and
               Ethan Dyer and
               Henryk Michalewski and
               Vinay V. Ramasesh and
               Ambrose Slone and
               Cem Anil and
               Imanol Schlag and
               Theo Gutman{-}Solo and
               Yuhuai Wu and
               Behnam Neyshabur and
               Guy Gur{-}Ari and
               Vedant Misra},
  editor    = {Sanmi Koyejo and
               S. Mohamed and
               A. Agarwal and
               Danielle Belgrave and
               K. Cho and
               A. Oh},
  title     = {Solving Quantitative Reasoning Problems with Language Models},
  booktitle = {Advances in Neural Information Processing Systems 35: Annual Conference
               on Neural Information Processing Systems 2022, NeurIPS 2022, New Orleans,
               LA, USA, November 28 - December 9, 2022},
  year      = {2022}
}

@inproceedings{olympiadbench,
  author    = {Chaoqun He and
               Renjie Luo and
               Yuzhuo Bai and
               Shengding Hu and
               Zhen Leng Thai and
               Junhao Shen and
               Jinyi Hu and
               Xu Han and
               Yujie Huang and
               Yuxiang Zhang and
               Jie Liu and
               Lei Qi and
               Zhiyuan Liu and
               Maosong Sun},
  editor    = {Lun{-}Wei Ku and
               Andre Martins and
               Vivek Srikumar},
  title     = {OlympiadBench: {A} Challenging Benchmark for Promoting {AGI} with
               Olympiad-Level Bilingual Multimodal Scientific Problems},
  booktitle = {Proceedings of the 62nd Annual Meeting of the Association for Computational
               Linguistics (Volume 1: Long Papers), {ACL} 2024, Bangkok, Thailand,
               August 11-16, 2024},
  pages     = {3828--3850},
  publisher = {Association for Computational Linguistics},
  year      = {2024}
}

@misc{aime2024,
  title        = {AIME 2024 (dataset card)},
  author       = {{HuggingFaceH4}},
  year         = {2024},
  howpublished = {Hugging Face},
  url          = {https://huggingface.co/datasets/HuggingFaceH4/aime_2024}
}

@misc{aime2025,
  title        = {AIME 2025 (dataset card)},
  author       = {{OpenCompass}},
  year         = {2025},
  howpublished = {Hugging Face},
  url          = {https://huggingface.co/datasets/opencompass/AIME2025}
}

@misc{amc23,
  title        = {AMC 2023 (dataset card)},
  author       = {{math-ai}},
  year         = {2025},
  howpublished = {Hugging Face},
  url          = {https://huggingface.co/datasets/math-ai/amc23}
}

@article{deepseekmath,
  title         = {DeepSeekMath: Pushing the Limits of Mathematical Reasoning in Open Language Models},
  author        = {Zhihong Shao and Peiyi Wang and Qihao Zhu and Runxin Xu and Junxiao Song and Xiao Bi and Haowei Zhang and Mingchuan Zhang and Y. K. Li and Y. Wu and Daya Guo},
  year          = {2024},
  eprint        = {2402.03300},
  archiveprefix = {arXiv},
  primaryclass  = {cs.CL},
  url           = {https://arxiv.org/abs/2402.03300}
}

@inproceedings{lightman2024letsverify,
  title     = {Let's Verify Step by Step},
  author    = {Hunter Lightman and Vineet Kosaraju and Yuri Burda and Harrison Edwards and Bowen Baker and Teddy Lee and Jan Leike and John Schulman and Ilya Sutskever and Karl Cobbe},
  booktitle = {The Twelfth International Conference on Learning Representations},
  year      = {2024}
}

@misc{math-verify,
  title        = {Math-verify: A robust mathematical expression evaluation system designed for assessing large language model outputs in mathematical tasks},
  author       = {{Hugging Face}},
  year         = {2025},
  howpublished = {GitHub repository},
  url          = {https://github.com/huggingface/Math-Verify}
}

@article{LLMJudgeSurvey2024,
  title   = {A survey on llm-as-a-judge},
  author  = {Jiawei Gu and Xuhui Jiang and Zhichao Shi and Hexiang Tan and Xuehao Zhai and Chengjin Xu and Wei Li and Yinghan Shen and Shengjie Ma and Honghao Liu and Saizhuo Wang and Kun Zhang and Yuanzhuo Wang and Wen Gao and Lionel Ni and Jian Guo},
  journal = {arXiv preprint arXiv:2411.15594},
  year    = {2024}
}

@article{williams1992reinforce,
  title     = {Simple statistical gradient-following algorithms for connectionist reinforcement learning},
  author    = {Williams, Ronald J},
  journal   = {Machine Learning},
  volume    = {8},
  number    = {3},
  pages     = {229--256},
  year      = {1992},
  publisher = {Springer}
}

@article{mroueh2025grpo,
  title   = {Reinforcement Learning with Verifiable Rewards: GRPO's Effective Loss, Dynamics, and Success Amplification},
  author  = {Mroueh, Youssef},
  journal = {arXiv preprint arXiv:2503.06639},
  year    = {2025}
}

@inproceedings{Wei2022CoT,
  author    = {Jason Wei and
               Xuezhi Wang and
               Dale Schuurmans and
               Maarten Bosma and
               Brian Ichter and
               Fei Xia and
               Ed H. Chi and
               Quoc V. Le and
               Denny Zhou},
  editor    = {Sanmi Koyejo and
               S. Mohamed and
               A. Agarwal and
               Danielle Belgrave and
               K. Cho and
               A. Oh},
  title     = {Chain-of-Thought Prompting Elicits Reasoning in Large Language Models},
  booktitle = {Advances in Neural Information Processing Systems 35: Annual Conference
               on Neural Information Processing Systems 2022, NeurIPS 2022, New Orleans,
               LA, USA, November 28 - December 9},
  year      = {2022}
}

@inproceedings{Wang2022SelfConsistency,
  author    = {Xuezhi Wang and
               Jason Wei and
               Dale Schuurmans and
               Quoc V. Le and
               Ed H. Chi and
               Sharan Narang and
               Aakanksha Chowdhery and
               Denny Zhou},
  title     = {Self-Consistency Improves Chain of Thought Reasoning in Language Models},
  booktitle = {The Eleventh International Conference on Learning Representations,
               {ICLR} 2023, Kigali, Rwanda, May 1-5, 2023},
  publisher = {OpenReview.net},
  year      = {2023}
}

@inproceedings{Zhou2022LeastToMost,
  author    = {Denny Zhou and
               Nathanael Sch{\"{a}}rli and
               Le Hou and
               Jason Wei and
               Nathan Scales and
               Xuezhi Wang and
               Dale Schuurmans and
               Claire Cui and
               Olivier Bousquet and
               Quoc V. Le and
               Ed H. Chi},
  title     = {Least-to-Most Prompting Enables Complex Reasoning in Large Language
               Models},
  booktitle = {The Eleventh International Conference on Learning Representations,
               {ICLR} 2023, Kigali, Rwanda, May 1-5, 2023},
  publisher = {OpenReview.net},
  year      = {2023}
}

@inproceedings{Yao2023ToT,
  author    = {Shunyu Yao and
               Dian Yu and
               Jeffrey Zhao and
               Izhak Shafran and
               Tom Griffiths and
               Yuan Cao and
               Karthik Narasimhan},
  editor    = {Alice Oh and
               Tristan Naumann and
               Amir Globerson and
               Kate Saenko and
               Moritz Hardt and
               Sergey Levine},
  title     = {Tree of Thoughts: Deliberate Problem Solving with Large Language Models},
  booktitle = {Advances in Neural Information Processing Systems 36: Annual Conference
               on Neural Information Processing Systems 2023, NeurIPS 2023, New Orleans,
               LA, USA, December 10 - 16},
  year      = {2023}
}

@article{Cobbe2021GSM8K,
  title   = {Training verifiers to solve math word problems},
  author  = {Karl Cobbe and
             Vineet Kosaraju and
             Mohammad Bavarian and
             Mark Chen and
             Heewoo Jun and
             Lukasz Kaiser and
             Matthias Plappert and
             Jerry Tworek and
             Jacob Hilton and
             Reiichiro Nakano and
             Christopher Hesse and
             John Schulman},
  journal = {arXiv preprint arXiv:2110.14168},
  year    = {2021}
}

@article{Thakur2024JudgingJudges,
  title   = {Judging the judges: Evaluating alignment and vulnerabilities in llms-as-judges},
  author  = {Thakur, Aman Singh and Choudhary, Kartik and Ramayapally, Venkat Srinik and Vaidyanathan, Sankaran and Hupkes, Dieuwke},
  journal = {arXiv preprint arXiv:2406.12624},
  year    = {2024}
}

@article{Li2025VerifyBench,
  title   = {VerifyBench: A Systematic Benchmark for Evaluating Reasoning Verifiers Across Domains},
  author  = {Li, Xuzhao and Li, Xuchen and Hu, Shiyu and Guo, Yongzhen and Zhang, Wentao},
  journal = {arXiv preprint arXiv:2507.09884},
  year    = {2025}
}

@article{song2020survey,
  title     = {Learning from noisy labels with deep neural networks: A survey},
  author    = {Song, Hwanjun and Kim, Minseok and Park, Dongmin and Shin, Yooju and Lee, Jae-Gil},
  journal   = {IEEE transactions on neural networks and learning systems},
  volume    = {34},
  number    = {11},
  pages     = {8135--8153},
  year      = {2022},
  publisher = {IEEE}
}

@inproceedings{natarajan2013,
  author    = {Nagarajan Natarajan and
               Inderjit S. Dhillon and
               Pradeep Ravikumar and
               Ambuj Tewari},
  editor    = {Christopher J. C. Burges and
               L{\'{e}}on Bottou and
               Zoubin Ghahramani and
               Kilian Q. Weinberger},
  title     = {Learning with Noisy Labels},
  booktitle = {Advances in Neural Information Processing Systems 26: 27th Annual
               Conference on Neural Information Processing Systems 2013. Proceedings
               of a meeting held December 5-8, 2013, Lake Tahoe, Nevada, United States},
  pages     = {1196--1204},
  year      = {2013}
}

@inproceedings{patrini2017,
  author    = {Giorgio Patrini and
               Alessandro Rozza and
               Aditya Krishna Menon and
               Richard Nock and
               Lizhen Qu},
  title     = {Making Deep Neural Networks Robust to Label Noise: {A} Loss Correction
               Approach},
  booktitle = {2017 {IEEE} Conference on Computer Vision and Pattern Recognition,
               {CVPR} 2017, Honolulu, HI, USA, July 21-26, 2017},
  pages     = {2233--2241},
  publisher = {{IEEE} Computer Society},
  year      = {2017}
}

@inproceedings{jiang2018mentornet,
  title        = {Mentornet: Learning data-driven curriculum for very deep neural networks on corrupted labels},
  author       = {Jiang, Lu and Zhou, Zhengyuan and Leung, Thomas and Li, Li-Jia and Fei-Fei, Li},
  booktitle    = {International conference on machine learning},
  pages        = {2304--2313},
  year         = {2018},
  organization = {PMLR}
}

@article{han2018coteaching,
  title   = {Co-teaching: Robust training of deep neural networks with extremely noisy labels},
  author  = {Han, Bo and Yao, Quanming and Yu, Xingrui and Niu, Gang and Xu, Miao and Hu, Weihua and Tsang, Ivor and Sugiyama, Masashi},
  journal = {Advances in neural information processing systems},
  volume  = {31},
  year    = {2018}
}

@inproceedings{li2020dividemix,
  author    = {Junnan Li and
               Richard Socher and
               Steven C. H. Hoi},
  title     = {DivideMix: Learning with Noisy Labels as Semi-supervised Learning},
  booktitle = {8th International Conference on Learning Representations, {ICLR} 2020,
               Addis Ababa, Ethiopia, April 26-30, 2020},
  publisher = {OpenReview.net},
  year      = {2020}
}

@article{zhao2025onetoken,
  title={One token to fool llm-as-a-judge},
  author={Zhao, Yulai and Liu, Haolin and Yu, Dian and Kung, Sunyuan and Chen, Meijia and Mi, Haitao and Yu, Dong},
  journal={arXiv preprint arXiv:2507.08794},
  year={2025}
}

@inproceedings{shi2025positionbias,
  title     = {Judging the judges: A systematic study of position bias in llm-as-a-judge},
  author    = {Shi, Lin and Ma, Chiyu and Liang, Wenhua and Diao, Xingjian and Ma, Weicheng and Vosoughi, Soroush},
  booktitle = {Proceedings of the 14th International Joint Conference on Natural Language Processing and the 4th Conference of the Asia-Pacific Chapter of the Association for Computational Linguistics},
  pages     = {292--314},
  year      = {2025}
}

@inproceedings{shi2024judgedeceiver,
  author    = {Jiawen Shi and
               Zenghui Yuan and
               Yinuo Liu and
               Yue Huang and
               Pan Zhou and
               Lichao Sun and
               Neil Zhenqiang Gong},
  editor    = {Bo Luo and
               Xiaojing Liao and
               Jun Xu and
               Engin Kirda and
               David Lie},
  title     = {Optimization-based Prompt Injection Attack to LLM-as-a-Judge},
  booktitle = {Proceedings of the 2024 on {ACM} {SIGSAC} Conference on Computer and
               Communications Security, {CCS} 2024, Salt Lake City, UT, USA, October
               14-18, 2024},
  pages     = {660--674},
  publisher = {{ACM}},
  year      = {2024}
}

@inproceedings{chen2024humansjudge,
  title     = {Humans or LLMs as the Judge? A Study on Judgement Bias},
  author    = {Chen, Guiming Hardy and Chen, Shunian and Liu, Ziche and Jiang, Feng and Wang, Benyou},
  booktitle = {Proceedings of the 2024 Conference on Empirical Methods in Natural Language Processing},
  pages     = {8301--8327},
  year      = {2024}
}

@article{huang2025pitfalls,
  title   = {Pitfalls of Rule- and Model-based Verifiers--A Case Study on Mathematical Reasoning},
  author  = {Huang, Yuzhen and Zeng, Weihao and Zeng, Xingshan and Zhu, Qi and He, Junxian},
  journal = {arXiv preprint arXiv:2505.22203},
  year    = {2025}
}

@inproceedings{sutton1999policy,
  author    = {Richard S. Sutton and
               David A. McAllester and
               Satinder Singh and
               Yishay Mansour},
  editor    = {Sara A. Solla and
               Todd K. Leen and
               Klaus{-}Robert M{\"{u}}ller},
  title     = {Policy Gradient Methods for Reinforcement Learning with Function Approximation},
  booktitle = {Advances in Neural Information Processing Systems 12, {[NIPS} Conference,
               Denver, Colorado, USA, November 29 - December 4, 1999]},
  pages     = {1057--1063},
  publisher = {The {MIT} Press},
  year      = {1999}
}

@inproceedings{wang2020reinforcement,
  author    = {Jingkang Wang and
               Yang Liu and
               Bo Li},
  title     = {Reinforcement Learning with Perturbed Rewards},
  booktitle = {The Thirty-Fourth {AAAI} Conference on Artificial Intelligence, {AAAI}
               2020, The Thirty-Second Innovative Applications of Artificial Intelligence
               Conference, {IAAI} 2020, The Tenth {AAAI} Symposium on Educational
               Advances in Artificial Intelligence, {EAAI} 2020, New York, NY, USA,
               February 7-12, 2020},
  pages     = {6202--6209},
  publisher = {{AAAI} Press},
  year      = {2020}
}

@inproceedings{hendrycks2021measuring,
  title={Measuring Massive Multitask Language Understanding},
  author={Dan Hendrycks and Collin Burns and Steven Basart and Andy Zou and Mantas Mazeika and Dawn Song and Jacob Steinhardt},
  booktitle={International Conference on Learning Representations},
  year={2021}
}

@article{clark2018think,
  title={Think you have Solved Question Answering? Try ARC, the AI2 Reasoning Challenge}, 
  author={Peter Clark and Isaac Cowhey and Oren Etzioni and Tushar Khot and Ashish Sabharwal and Carissa Schoenick and Oyvind Tafjord},
  year={2018},
  eprint={1803.05457},
  archivePrefix={arXiv},
  primaryClass={cs.AI}
}

@inproceedings{zellers2019hellaswag,
  title={Hellaswag: Can a machine really finish your sentence?},
  author={Zellers, Rowan and Holtzman, Ari and Bisk, Yonatan and Farhadi, Ali and Choi, Yejin},
  booktitle={Proceedings of the 57th annual meeting of the association for computational linguistics},
  pages={4791--4800},
  year={2019}
}

@inproceedings{rein2023gpqa,
  title={{GPQA}: A Graduate-Level Google-Proof Q\&A Benchmark},
  author={David Rein and Betty Li Hou and Asa Cooper Stickland and Jackson Petty and Richard Yuanzhe Pang and Julien Dirani and Julian Michael and Samuel R. Bowman},
  booktitle={First Conference on Language Modeling},
  year={2024}
}

@misc{chen2021evaluatinglargelanguagemodels,
      title={Evaluating Large Language Models Trained on Code}, 
      author={Mark Chen and Jerry Tworek and Heewoo Jun and Qiming Yuan and Henrique Ponde de Oliveira Pinto and Jared Kaplan and Harri Edwards and Yuri Burda and Nicholas Joseph and Greg Brockman and Alex Ray and Raul Puri and Gretchen Krueger and Michael Petrov and Heidy Khlaaf and Girish Sastry and Pamela Mishkin and Brooke Chan and Scott Gray and Nick Ryder and Mikhail Pavlov and Alethea Power and Lukasz Kaiser and Mohammad Bavarian and Clemens Winter and Philippe Tillet and Felipe Petroski Such and Dave Cummings and Matthias Plappert and Fotios Chantzis and Elizabeth Barnes and Ariel Herbert-Voss and William Hebgen Guss and Alex Nichol and Alex Paino and Nikolas Tezak and Jie Tang and Igor Babuschkin and Suchir Balaji and Shantanu Jain and William Saunders and Christopher Hesse and Andrew N. Carr and Jan Leike and Josh Achiam and Vedant Misra and Evan Morikawa and Alec Radford and Matthew Knight and Miles Brundage and Mira Murati and Katie Mayer and Peter Welinder and Bob McGrew and Dario Amodei and Sam McCandlish and Ilya Sutskever and Wojciech Zaremba},
      year={2021},
      eprint={2107.03374},
      archivePrefix={arXiv},
      primaryClass={cs.LG},
      url={https://arxiv.org/abs/2107.03374}, 
}

@inproceedings{liu-etal-2025-compassverifier,
    title = "{C}ompass{V}erifier: A Unified and Robust Verifier for {LLM}s Evaluation and Outcome Reward",
    author = "Liu, Shudong  and
      Liu, Hongwei  and
      Liu, Junnan  and
      Xiao, Linchen  and
      Gao, Songyang  and
      Lyu, Chengqi  and
      Gu, Yuzhe  and
      Zhang, Wenwei  and
      Wong, Derek F.  and
      Zhang, Songyang  and
      Chen, Kai",
    editor = "Christodoulopoulos, Christos  and
      Chakraborty, Tanmoy  and
      Rose, Carolyn  and
      Peng, Violet",
    booktitle = "Proceedings of the 2025 Conference on Empirical Methods in Natural Language Processing",
    month = nov,
    year = "2025",
    address = "Suzhou, China",
    publisher = "Association for Computational Linguistics",
    url = "https://aclanthology.org/2025.emnlp-main.1698/",
    doi = "10.18653/v1/2025.emnlp-main.1698",
    pages = "33466--33494",
    ISBN = "979-8-89176-332-6"
}

\newpage
\appendix

\section{From REINFORCE-Style Corrections to GRPO Updates}
\label{app:REINFORCE-to-GRPO}
\paragraph{Setup (single prompt and group).}
Fix a prompt $x$ and sample a group of $K$ i.i.d.\ responses $y_1,\dots,y_K \sim \pi_\theta(\cdot\mid x)$.
Let $G_i \triangleq \sum_{t=1}^{|y_i|}\nabla_\theta \log \pi_\theta(y_{i,t}\mid x,y_{i,<t})$ denote the trajectory score-function.
Let $u_i$ be any scalar \emph{proxy reward} computed from the trajectory. For PGBC, $u_i=\widehat r_i$. For the population PGFC theorem, the score-function term is weighted as $w_{\tilde r_i}G_i$; the hypothetical pre-centering proxy $u_i=\tilde r_i w_{\tilde r_i}$ is analyzed below only as a degeneracy example, not as our default implementation.
Define the group mean $\bar u = \frac{1}{K}\sum_{j=1}^K u_j$.
Consider the centered score-function update
\begin{equation}
\label{eq:centered-update}
\Delta_{\text{center}}(\theta)\;\triangleq\;\frac{1}{K}\sum_{i=1}^K (u_i-\bar u)\,G_i.
\end{equation}

\begin{proposition}[Group centering preserves expected direction]
\label{prop:centering-preserves-direction}
Assume $\{(u_i,G_i)\}_{i=1}^K$ are i.i.d.\ under $\pi_\theta(\cdot\mid x)$ and $\E[G_i\mid x]=0$ (the standard score-function identity~\citep{williams1992reinforce,sutton1999policy}).
Then the centered update in~\eqref{eq:centered-update} satisfies
\begin{equation}
\label{eq:centering-scale}
\E\!\left[\Delta_{\text{center}}(\theta)\mid x\right]\;=\;\Bigl(1-\frac{1}{K}\Bigr)\,\E\!\left[u\,G\mid x\right].
\end{equation}
In particular, if $\E[u\,G\mid x]$ is parallel to the clean gradient direction for the prompt-conditioned objective, then so is $\E[\Delta_{\text{center}}(\theta)\mid x]$.
\end{proposition}

\begin{proof}
Expand $\Delta_{\text{center}}=\frac1K\sum_i u_iG_i - \bar u\cdot\frac1K\sum_i G_i$.
Taking expectation conditioned on $x$ yields
\[
\E[\bar u\cdot \tfrac1K\sum_i G_i\mid x]
=\frac{1}{K^2}\sum_{i,j}\E[u_j G_i\mid x].
\]
The $i=j$ terms contribute $\frac{K}{K^2}\E[uG\mid x]$.
For $i\neq j$, independence gives $\E[u_j G_i\mid x]=\E[u\mid x]\E[G\mid x]=0$.
Thus $\E[\bar u\cdot \tfrac1K\sum_i G_i\mid x]=\frac{1}{K}\E[uG\mid x]$, implying~\eqref{eq:centering-scale}.
\end{proof}

\begin{corollary}[Directional correctness of PGFC under centered GRPO-style updates]
\label{cor:pgfc-centered-grpo}
Under the Verifier Reward Channel (Definition~3.1) and the conditional-independence assumption stated above, let $u=w_{\tilde R}$ with $w_0=\rho_1-1$ and $w_1=\rho_1$ as in Proposition~3.4.
Then for each prompt $x$,
\[
\E\!\left[\Delta_{\text{center}}(\theta)\mid x\right]
\;=\;\Bigl(1-\frac{1}{K}\Bigr)(1-\rho_0-\rho_1)\,\nabla_\theta J(\theta\mid x),
\]
so the expected centered update remains aligned with the clean policy-gradient direction.
\end{corollary}

\paragraph{Binary-proxy equivalence caveat.}
The corollary explains how centering interacts with a generic proxy whose score-function expectation is aligned with the clean gradient. It does not by itself prove that every PGFC proxy creates a distinct GRPO update. For binary rewards, \(w_{\tilde r}=\tilde r+\rho_1-1\) differs from \(\tilde r\) by an additive constant, so group centering removes the difference. Likewise, the hypothetical pre-centering proxy \(u=\tilde r\,w_{\tilde r}=\rho_1\tilde r\) is a positive rescaling of \(\tilde r\); if within-group standardization is enabled, this rescaling also cancels.

\paragraph{Standardization, clipping, and KL (what is guaranteed, and what is approximate).}
GRPO further divides centered proxy rewards by the within-group standard deviation to form normalized advantages, and then applies a PPO-style clipping on token-level ratios and an explicit KL penalty to $\pi_{\mathrm{ref}}$~\citep{deepseekmath}.
The exact direction-preservation statement in Proposition~\ref{prop:centering-preserves-direction} covers the \emph{centering} step, which is the main nontrivial departure from vanilla REINFORCE.
Clipping and the KL penalty add further stabilizers: in the small-update regime (encouraged by the KL term and typical learning rates), clipping is rarely active and the update is close to the unclipped score-function form; when clipping activates, it intentionally limits step size and may deviate from the exact REINFORCE direction.
Importantly, our corrections target the \emph{reward-gradient component}; the KL gradient is unchanged, so any global scaling factor (e.g., $1-\rho_0-\rho_1$) can be absorbed into the effective learning rate and/or interpreted as modifying the reward--KL trade-off. 

\section{Implementation Details}
\label{app:impl-grpo}

This appendix describes the implementation of PGBC and PGFC inside a GRPO-style RLVR pipeline.
The key design choice is to insert each correction at the level of the scalar coefficient multiplying the score-function term.
PGBC is a \emph{corrected-utility} method: it replaces the observed verifier reward by an estimate of the clean reward.
PGFC is a \emph{score-corrected} method: it replaces the usual reward-derived advantage by a signed channel-calibrated multiplier.
In both cases, the corrected coefficient is passed directly to the PPO-style GRPO surrogate and is not converted into a binary reward proxy that is subsequently group-standardized.

\subsection{Notation and GRPO Surrogate}

For a prompt \(x\), GRPO samples a group of \(K\) responses
\[
y_1,\ldots,y_K\sim \pi_{\theta_{\mathrm{old}}}(\cdot\mid x).
\]
Let \(\tilde r_i\in\{0,1\}\) be the observed verifier reward for response \(y_i\).
For token \(t\) in response \(i\), define the policy ratio
\[
r_{i,t}(\theta)
=
\frac{\pi_\theta(y_{i,t}\mid x,y_{i,<t})}
     {\pi_{\theta_{\mathrm{old}}}(y_{i,t}\mid x,y_{i,<t})}.
\]
Given a scalar trajectory coefficient \(a_i\), we broadcast it to all response tokens:
\[
A_{i,t}\equiv a_i,\qquad t=1,\ldots,|y_i|.
\]
The clipped GRPO reward surrogate is then
\[
\mathcal L_{\mathrm{GRPO}}^{\mathrm{rew}}(\theta)
=
\frac{1}{K}\sum_{i=1}^K
\frac{1}{|y_i|}
\sum_{t=1}^{|y_i|}
\min\!\left(
r_{i,t}(\theta) A_{i,t},
\operatorname{clip}(r_{i,t}(\theta),1-\varepsilon,1+\varepsilon) A_{i,t}
\right).
\]
The full objective keeps the usual KL penalty:
\[
\mathcal L_{\mathrm{GRPO}}(\theta)
=
\mathcal L_{\mathrm{GRPO}}^{\mathrm{rew}}(\theta)
-
\beta\,
\mathrm{KL}\!\left(\pi_\theta(\cdot\mid x)\,\|\,\pi_{\mathrm{ref}}(\cdot\mid x)\right).
\]
At \(\theta=\theta_{\mathrm{old}}\), when clipping is inactive, the reward-gradient component reduces to the score-function update
\[
\nabla_\theta \mathcal L_{\mathrm{GRPO}}^{\mathrm{rew}}(\theta)
=
\frac{1}{K}\sum_{i=1}^K a_i
\left(
\frac{1}{|y_i|}\sum_{t=1}^{|y_i|}
\nabla_\theta\log\pi_\theta(y_{i,t}\mid x,y_{i,<t})
\right),
\]
up to the conventional length normalization used by the implementation.
Therefore, the scalar coefficient \(a_i\) is the correct place to insert PGBC or PGFC.

\subsection{PGBC: Corrected-Utility GRPO}

PGBC estimates the clean binary reward by inverting the verifier channel.
Given estimates \(\hat\rho_0,\hat\rho_1\), define
\[
\hat c = 1-\hat\rho_0-\hat\rho_1.
\]
We require \(\hat c>0\) and in code clamp it below by a small constant \(\epsilon_c>0\) for numerical safety.
The corrected utility is
\[
\widehat r_i
=
\frac{\tilde r_i-\hat\rho_0}{\hat c}.
\]
This quantity is used as the trajectory coefficient, optionally with a variance-reduction baseline:
\[
a_i^{\mathrm{BC}}
=
\frac{\widehat r_i-b_i^{\mathrm{BC}}}{s_{\mathrm{BC}}}.
\]
Here \(s_{\mathrm{BC}}>0\) is a fixed or slowly moving positive scale, such as an EMA scale, and should not be the current group's standard deviation.
The default baseline is leave-one-out:
\[
b_i^{\mathrm{BC}}
=
\frac{1}{K-1}\sum_{j\neq i}\widehat r_j
\qquad (K>1),
\]
with \(b_i^{\mathrm{BC}}=0\) when \(K=1\).
The leave-one-out baseline is independent of \(G_i\) conditional on \(x\), so before clipping and KL it preserves the expected score-function update:
\[
\mathbb E\!\left[
\frac{1}{K}\sum_{i=1}^K
(\widehat r_i-b_i^{\mathrm{BC}})G_i
\,\middle|\,x
\right]
=
\mathbb E[\widehat r\,G\mid x].
\]
When the channel rates are correct, \(\mathbb E[\widehat r\,G\mid x]\) equals the clean reward-gradient component.
Using the group mean including the \(i\)-th sample is also possible, but it introduces the familiar multiplicative factor \(1-1/K\).

The corrected coefficient \(a_i^{\mathrm{BC}}\) is then broadcast to all tokens:
\[
A_{i,t}^{\mathrm{BC}}\equiv a_i^{\mathrm{BC}}.
\]
No current-group \(z\)-score standardization is applied to \(\{\widehat r_i\}_{i=1}^K\), because such standardization would remove the affine calibration induced by PGBC in the binary-reward case.

\subsection{PGFC: Score-Corrected GRPO}

PGFC is implemented as a score-level correction rather than as a reward proxy.
Given an estimate \(\hat\rho_1\), define the signed score multiplier
\[
w_{\tilde r_i}
=
\tilde r_i-(1-\hat\rho_1)
=
\begin{cases}
\hat\rho_1-1, & \tilde r_i=0,\\
\hat\rho_1, & \tilde r_i=1.
\end{cases}
\]
The trajectory coefficient is
\[
a_i^{\mathrm{FC}}
=
\frac{w_{\tilde r_i}-b_i^{\mathrm{FC}}}{s_{\mathrm{FC}}}.
\]
Our default is
\[
b_i^{\mathrm{FC}}=0.
\]
The reason is that subtracting the current group mean of \(w_{\tilde r_i}\) would reduce PGFC to an ordinary relative noisy-reward update, because \(w_{\tilde r_i}\) is an affine function of \(\tilde r_i\).
A fixed or historical baseline independent of the current sampled actions may be used, but current-group centering is not the default.

The scale \(s_{\mathrm{FC}}>0\) controls the reward--KL trade-off.
For pure score-function direction, any fixed positive scale preserves the direction.
When both noise rates are available, one may set
\[
s_{\mathrm{FC}} = 1-\hat\rho_0-\hat\rho_1
\]
to approximately match the clean reward-gradient magnitude.
In the FN-dominant rule-checker regime where \(\rho_0\approx 0\), a practical choice is
\[
s_{\mathrm{FC}} = 1-\hat\rho_1.
\]
If no calibrated scale is desired, we set \(s_{\mathrm{FC}}=1\) and tune \(\beta\) or the learning rate as usual.

The token-level advantage is
\[
A_{i,t}^{\mathrm{FC}}\equiv a_i^{\mathrm{FC}}.
\]
Before clipping and KL, if \(\hat\rho_1=\rho_1\), the population score-function component satisfies
\[
\mathbb E[a_i^{\mathrm{FC}}G_i\mid x]
=
\frac{1-\rho_0-\rho_1}{s_{\mathrm{FC}}}
\nabla_\theta J(\theta\mid x).
\]
Thus PGFC preserves the clean reward-gradient direction and, with a suitable positive scale, can also calibrate the reward-gradient magnitude.
The KL gradient is not corrected; therefore \(s_{\mathrm{FC}}\) and \(\beta\) jointly determine the effective reward--KL trade-off.

\paragraph{Remark on the role of \(\hat\rho_1\).}
The theoretical conditional-cancellation identity for PGFC uses the true \(\rho_1\).
In an unclipped population score-function estimator, adding any action-independent constant to \(\tilde r_i\) has zero expected gradient by the score-function identity.
However, in finite-batch GRPO with clipping, KL regularization, length normalization, and nonstationary rollouts, the fixed channel baseline \(1-\hat\rho_1\) affects the realized signed advantages and the clipped surrogate.

\subsection{Why We Do Not Use Direct-Group \(z\)-Scoring for Binary Channel Corrections}

Let \(Z_i(u)\) denote the standard group-normalized advantage
\[
Z_i(u)
=
\frac{u_i-\bar u}{\operatorname{std}(\{u_j\}_{j=1}^K)+\epsilon},
\qquad
\bar u=\frac{1}{K}\sum_{j=1}^K u_j.
\]
If \(u_i=a\tilde r_i+b\) with \(a>0\), then
\[
u_i-\bar u
=
a(\tilde r_i-\bar{\tilde r}),
\qquad
\operatorname{std}(\{u_j\})
=
a\,\operatorname{std}(\{\tilde r_j\}),
\]
and therefore
\[
Z_i(u)
=
\frac{\tilde r_i-\bar{\tilde r}}
     {\operatorname{std}(\{\tilde r_j\})+\epsilon/a}.
\]
When \(\epsilon\) is small and the group is non-degenerate, this is numerically identical to the uncorrected noisy-reward advantage.

This degeneracy applies to the most obvious binary implementations of both corrections:
\[
\widehat r_i
=
\frac{\tilde r_i-\hat\rho_0}{1-\hat\rho_0-\hat\rho_1}
\quad\text{is affine in }\tilde r_i,
\]
and
\[
w_{\tilde r_i}
=
\tilde r_i+\hat\rho_1-1,
\qquad
\tilde r_i w_{\tilde r_i}
=
\hat\rho_1\tilde r_i
\quad\text{for }\tilde r_i\in\{0,1\}.
\]
Thus, feeding any of these quantities into the usual current-group \(z\)-score advantage construction can make the corrected method exactly equivalent to the uncorrected noisy baseline.
For this reason, our corrected implementations use the scalar coefficients \(a_i^{\mathrm{BC}}\) and \(a_i^{\mathrm{FC}}\) directly, with only fixed or running positive scales.

\subsection{Process-Supervised Extension}

For process supervision, suppose response \(y_i\) has step-level verifier rewards
\[
\tilde r_i^{(1)},\ldots,\tilde r_i^{(J_i)}
\]
attached to step boundaries
\[
\tau_i^{(1)}<\cdots<\tau_i^{(J_i)}.
\]
PGBC is applied to each step reward:
\[
\widehat r_i^{(j)}
=
\frac{\tilde r_i^{(j)}-\hat\rho_0}
     {1-\hat\rho_0-\hat\rho_1}.
\]
The token coefficient is the return-to-go of corrected utilities:
\[
A_{i,t}^{\mathrm{BC}}
=
\frac{1}{s_{\mathrm{BC}}}
\sum_{j:\tau_i^{(j)}\ge t}
\left(\widehat r_i^{(j)}-b_{i,j}^{\mathrm{BC}}\right),
\]
where \(b_{i,j}^{\mathrm{BC}}\) is an optional leave-one-out or historical baseline for the corresponding step.

PGFC is applied as a step-level signed score multiplier:
\[
w_i^{(j)}
=
\tilde r_i^{(j)}-(1-\hat\rho_1),
\]
and the token coefficient is
\[
A_{i,t}^{\mathrm{FC}}
=
\frac{1}{s_{\mathrm{FC}}}
\sum_{j:\tau_i^{(j)}\ge t}
w_i^{(j)}.
\]
As in the outcome case, we avoid current-batch \(z\)-scoring of these binary affine quantities.
If a positive normalization is needed for numerical stability, we use a fixed or EMA scale rather than the current batch's standard deviation.

\subsection{When Conditional Independence Is Violated (Bias Decomposition).}
Our directional-alignment results rely on conditional independence to factorize
$\E[w_{\tilde R}G\mid R^*]=\E[w_{\tilde R}\mid R^*]\E[G\mid R^*]$.
If verifier errors depend on trajectory features beyond correctness (e.g., style-based FP hacks), the update admits the decomposition
\[
\E[w_{\tilde R}G\mid R^*] \;=\; \E[w_{\tilde R}\mid R^*]\E[G\mid R^*] \;+\; \mathrm{Cov}(w_{\tilde R},G\mid R^*),
\]
so the residual bias is governed by the conditional covariance term.
This highlights a practical diagnostic: if verifier decisions are strongly predictable from superficial cues that also affect generation likelihood, $\mathrm{Cov}(w_{\tilde R},G\mid R^*)$ may be non-negligible and one should combine channel-based correction with verifier hardening or instance-dependent reliability modeling.

\subsection{Online FN-Rate Estimation via Appeals}

In the FN-dominant rule-checker regime, PGFC requires an estimate of the false-negative rate \(\rho_1\).
At training step \(t\), let
\[
\mathcal N_{\mathrm R}^{(t)}
=
\{i:\tilde r_i=0\},
\qquad
\mathcal P_{\mathrm R}^{(t)}
=
\{i:\tilde r_i=1\}
\]
be the rule-based negatives and positives.
We appeal each rule-negative item independently with probability \(q\in(0,1]\) using a lightweight auxiliary verifier.
Let \(S^{(t)}\subseteq\mathcal N_{\mathrm R}^{(t)}\) be the appealed subset, and let
\[
\mathcal P_{\mathrm L}^{(t)}
\subseteq S^{(t)}
\]
be the appealed items flipped to positive by the auxiliary verifier.
The Horvitz--Thompson estimate of the number of false negatives is
\[
\widehat{\mathrm{FN}}^{(t)}
=
\frac{|\mathcal P_{\mathrm L}^{(t)}|}{q}.
\]
Assuming rule-based positives have negligible FP rate, we estimate
\[
\hat\rho_1^{(t)}
=
\frac{\widehat{\mathrm{FN}}^{(t)}+\alpha}
     {\widehat{\mathrm{FN}}^{(t)}
      +|\mathcal P_{\mathrm R}^{(t)}|
      +\alpha+\beta},
\]
where \(\alpha,\beta>0\) are small pseudocounts.
To track policy drift, we use EMA smoothing:
\[
\bar\rho_1^{(t)}
=
\lambda\bar\rho_1^{(t-1)}
+
(1-\lambda)\hat\rho_1^{(t)}.
\]
The smoothed value \(\bar\rho_1^{(t)}\) is used as \(\hat\rho_1\) in PGFC.
The expected number of auxiliary verifier calls per update is
\[
q\,|\mathcal N_{\mathrm R}^{(t)}|.
\]

For PGBC, both \(\rho_0\) and \(\rho_1\) are needed.
When \(\rho_0\) is unavailable or unreliable, we prefer PGFC in the FN-dominant setting.
When both rates can be estimated from a calibration set or from a dual-verifier protocol, PGBC can be used with the corrected utility above.

\section{Additional Experimental Results}
\label{app:additional-experiments}

\subsection{Performance under Pass@8 Metric}
Table~\ref{tab:realworld-all-pass8} retains the measured pass@8 results for the four backbones. The conclusion remains consistent with the pass@1 results: PGBC and PGFC provide reliable improvements over the noisy baseline.

\subsection{Complementarity with Stronger Verifiers}
We additionally investigate whether the proposed correction mechanisms are synergistic with more capable verifiers. While our main experiments instantiated the appeal process with a lightweight LLM verifier (TinyV) to minimize computational overhead, Table~\ref{tab:stronger-checker} demonstrates that the forward correction yields consistent improvements even when complementing stronger verifier baselines. Specifically, we compare against a ``Stronger checker'' that extracts the final \verb|\boxed{}| answer and applies numeric/rational equivalence checking, using the \texttt{Math-Verify} library as a robust baseline. Concretely, we parse the gold answer using a gold-appropriate extraction configuration (\texttt{ExprExtractionConfig} for plain expressions or \texttt{LatexExtractionConfig} for LaTeX), and parse model predictions using \texttt{LatexExtractionConfig} together with \texttt{ExprExtractionConfig}. Verification is then performed by \texttt{Math-Verify}'s parse-and-verify pipeline with reward-verification-oriented normalization. Under this setup, PGFC uses the standard rule-based reward as the primary signal and calls the auxiliary \texttt{Math-Verify} checker only on appealed rule-negative samples. With the synthetic noise rates fixed at $\rho_0=0.1$ and $\rho_1=0.2$ as in the main paper, our \texttt{Appeals+PGFC} method achieves notable gains over the strong baseline (e.g., on MATH500 and AMC2023). These results suggest that the theoretical corrections act orthogonally to verifier engineering: regardless of how accurate the underlying verifier is, explicitly modeling and correcting for the residual false-negative rate provides a reliable performance lift.

\subsection{Training Dynamics under Synthetic Noise}
\begin{figure}[htbp]
  \centering
  \includegraphics[width=0.95\linewidth]{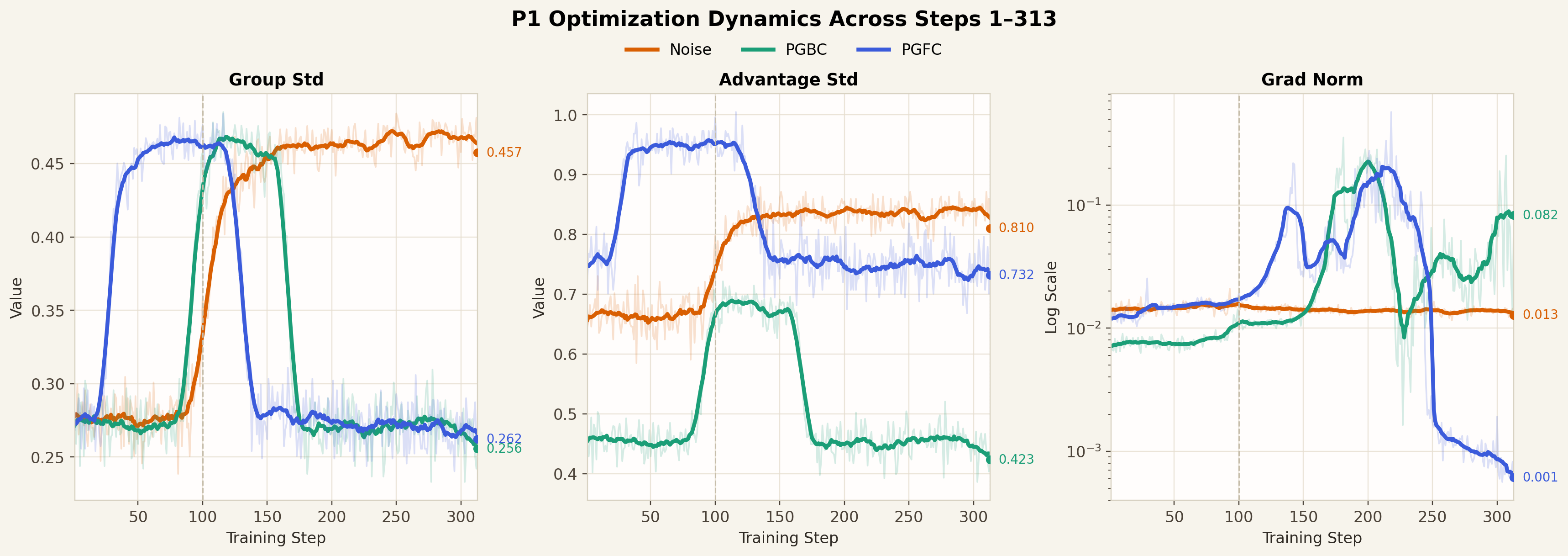}
  \caption{Training dynamics under synthetic iid noise. PGBC and PGFC change the group reward standard deviation, normalized advantage standard deviation, and gradient norm relative to uncorrected noisy training, illustrating that the correction is active inside advantage construction rather than only changing endpoint scores.}
  \label{fig:training-dynamics}
\end{figure}

To better understand how the proposed corrections influence the learning process, we analyze the training dynamics under synthetic i.i.d. noise. As shown in Figure~\ref{fig:training-dynamics}, we monitor the group reward standard deviation, normalized advantage standard deviation, and gradient norm across training steps. The purpose of this analysis is to verify that the backward and forward corrections actively modify the advantage construction during training, rather than merely shifting the final endpoint scores. The results demonstrate that both PGBC and PGFC measurably alter these intermediate training statistics relative to the uncorrected noisy baseline, providing evidence that our corrections dynamically adjust the policy gradient throughout the optimization process.

\begin{table*}[htbp]
  \centering
  \small
  \caption{Pass@8 under real-world noise.}
  \label{tab:realworld-all-pass8}
  \resizebox{\linewidth}{!}{
    \begin{tabular}{lrrrrrr|r}
      \toprule
      Dataset & AIME2024 & AIME2025 & AMC2023 & MATH500 & Minerva MATH & Olympiad Bench & Average \\
      \midrule
      \multicolumn{8}{c}{Qwen2.5-Math-1.5B} \\
Base & 32.4 $\pm$ 0.5 & 16.7 $\pm$ 0.7 & 79.3 $\pm$ 1.4 & 62.8 $\pm$ 1.3 & 15.5 $\pm$ 1.1 & 31.3 $\pm$ 0.8 & 39.7 $\pm$ 1.0 \\
Rule & \textbf{34.3 $\pm$ 1.2} & 19.9 $\pm$ 1.1 & 80.7 $\pm$ 1.4 & 69.6 $\pm$ 0.5 & 18.0 $\pm$ 0.6 & 32.8 $\pm$ 1.2 & \textbf{42.6 $\pm$ 1.0} \\
LLM-as-Judge & 29.6 $\pm$ 1.1 & 15.4 $\pm$ 0.4 & 80.0 $\pm$ 0.6 & 63.4 $\pm$ 1.2 & 16.2 $\pm$ 0.8 & 28.6 $\pm$ 0.3 & 38.9 $\pm$ 0.7 \\
Appeals & 30.5 $\pm$ 0.2 & \textbf{20.5 $\pm$ 0.7} & 80.7 $\pm$ 0.9 & 68.9 $\pm$ 1.3 & 17.6 $\pm$ 1.0 & 30.5 $\pm$ 0.5 & 41.5 $\pm$ 0.8 \\
Appeals+PGFC (Ours) & 31.0 $\pm$ 0.2 & 20.0 $\pm$ 0.6 & \textbf{82.2 $\pm$ 0.4} & \textbf{69.8 $\pm$ 1.1} & \textbf{18.2 $\pm$ 0.5} & \textbf{33.3 $\pm$ 0.3} & 42.4 $\pm$ 0.5 \\
      \midrule
      \multicolumn{8}{c}{DeepSeek-R1-Distill-Qwen-1.5B} \\
Base & 44.8 $\pm$ 0.8 & 32.8 $\pm$ 0.7 & 88.2 $\pm$ 0.1 & 85.9 $\pm$ 0.6 & 27.8 $\pm$ 0.3 & 48.6 $\pm$ 0.3 & 54.7 $\pm$ 0.4 \\
Rule & 55.1 $\pm$ 0.6 & 40.1 $\pm$ 0.4 & 96.1 $\pm$ 0.2 & 96.6 $\pm$ 0.1 & 44.6 $\pm$ 0.1 & 64.9 $\pm$ 0.1 & 66.2 $\pm$ 0.1 \\
LLM-as-Judge & 55.4 $\pm$ 0.9 & 39.8 $\pm$ 0.7 & 96.2 $\pm$ 0.2 & 95.7 $\pm$ 0.2 & 43.2 $\pm$ 0.7 & 62.8 $\pm$ 0.7 & 65.5 $\pm$ 0.3 \\
Appeals & 57.7 $\pm$ 1.0 & 42.5 $\pm$ 0.8 & 96.0 $\pm$ 0.2 & 96.4 $\pm$ 0.2 & 44.2 $\pm$ 0.8 & 64.5 $\pm$ 0.8 & 66.9 $\pm$ 0.3 \\
Appeals+PGFC (Ours) & \textbf{61.3 $\pm$ 0.6} & \textbf{44.6 $\pm$ 0.7} & \textbf{98.6 $\pm$ 0.1} & \textbf{97.1 $\pm$ 0.2} & \textbf{46.8 $\pm$ 0.6} & \textbf{66.1 $\pm$ 0.4} & \textbf{69.1 $\pm$ 0.2} \\
      \midrule
      \multicolumn{8}{c}{Llama-3.2-3B-Instruct} \\
Base & 18.8 $\pm$ 1.1 & 3.3 $\pm$ 0.2 & 54.7 $\pm$ 1.4 & 40.8 $\pm$ 0.8 & 5.2 $\pm$ 1.1 & 16.4 $\pm$ 0.5 & 23.2 $\pm$ 0.8 \\
Rule & \textbf{22.4 $\pm$ 1.0} & \textbf{6.7 $\pm$ 1.4} & \textbf{60.0 $\pm$ 1.3} & 44.8 $\pm$ 0.6 & \textbf{13.3 $\pm$ 0.7} & \textbf{18.7 $\pm$ 1.5} & \textbf{27.7 $\pm$ 1.1} \\
LLM-as-Judge & 10.0 $\pm$ 0.9 & 0.8 $\pm$ 0.9 & 44.4 $\pm$ 1.4 & 27.2 $\pm$ 0.9 & 6.7 $\pm$ 1.4 & 9.5 $\pm$ 0.2 & 16.4 $\pm$ 0.9 \\
Appeals & 17.5 $\pm$ 0.2 & 4.2 $\pm$ 1.0 & 53.8 $\pm$ 0.2 & 42.3 $\pm$ 1.5 & 10.8 $\pm$ 1.6 & 14.9 $\pm$ 1.2 & 23.9 $\pm$ 0.9 \\
Appeals+PGFC (Ours) & 20.8 $\pm$ 0.9 & 3.3 $\pm$ 1.5 & 58.1 $\pm$ 1.3 & \textbf{47.6 $\pm$ 0.6} & 12.6 $\pm$ 1.7 & 18.5 $\pm$ 0.6 & 26.8 $\pm$ 1.0 \\
      \midrule
      \multicolumn{8}{c}{Qwen2.5-Math-7B} \\
Base & 36.2 $\pm$ 1.2 & 23.3 $\pm$ 1.0 & 81.9 $\pm$ 1.3 & 66.6 $\pm$ 0.5 & 11.8 $\pm$ 0.6 & 33.9 $\pm$ 0.7 & 42.3 $\pm$ 0.9 \\
Rule & 50.0 $\pm$ 0.1 & 28.7 $\pm$ 0.8 & 83.4 $\pm$ 0.6 & 82.8 $\pm$ 0.5 & \textbf{29.0 $\pm$ 1.1} & 45.8 $\pm$ 1.3 & 53.3 $\pm$ 0.7 \\
LLM-as-Judge & 41.7 $\pm$ 0.3 & 22.1 $\pm$ 0.3 & 81.9 $\pm$ 1.5 & 69.0 $\pm$ 0.2 & 13.2 $\pm$ 1.2 & 39.4 $\pm$ 1.1 & 44.6 $\pm$ 0.8 \\
Appeals & 47.1 $\pm$ 1.0 & \textbf{30.4 $\pm$ 0.3} & \textbf{84.4 $\pm$ 1.2} & 80.8 $\pm$ 0.3 & 23.5 $\pm$ 1.3 & 45.6 $\pm$ 0.5 & 52.0 $\pm$ 0.8 \\
Appeals+PGFC (Ours) & \textbf{54.6 $\pm$ 1.4} & \textbf{30.4 $\pm$ 1.2} & 82.8 $\pm$ 0.2 & \textbf{83.2 $\pm$ 1.3} & \textbf{29.0 $\pm$ 0.6} & \textbf{47.6 $\pm$ 0.9} & \textbf{54.6 $\pm$ 0.9} \\
      \bottomrule
    \end{tabular}
  }
\end{table*}

\subsection{Appeals Statistics and Computational Overhead}
To quantify the computational cost and the empirical behavior of the appeal mechanism under real-world verifier noise, we track the appeal and flip statistics during training on \textit{Qwen2.5-Math-1.5B}. As detailed in Table~\ref{tab:appeals-stats}, we measure the \emph{appeal rate} (the fraction of rollout responses that trigger an appeal) and the \emph{flip rate} (the fraction of appealed responses that are successfully overturned by the LLM judge). The two methods use the same negative-side appeal mechanism. Within Appeals+PGFC, these calls support both reward recovery and online FN-rate estimation, without a separate post-appeal verifier pass. The total number of auxiliary \texttt{TinyV} calls remains small relative to the total number of training rollouts, demonstrating that our online false-negative rate estimation and forward correction are highly computationally efficient.

\begin{table}[htbp]
  \centering
  \small
  \caption{Appeals statistics for Qwen2.5-Math-1.5B real-world verifier noise. Appeals+PGFC uses the same appeal stream both for recovery and for online FN-rate estimation.}
  \label{tab:appeals-stats}
  \begin{tabular}{lccccc}
    \toprule
    Method & Avg pass@1 & Avg pass@8 & Appeal rate & Flip rate & TinyV calls \\
    \midrule
    Rule & $31.6{\pm}0.4$ & $\mathbf{42.6{\pm}1.0}$ & 0.000 & 0.0000 & 0 \\
    Appeals & $30.1{\pm}0.6$ & $41.5{\pm}0.8$ & 0.081 & 0.2288 & 7,963 \\
    Appeals+PGFC & $\mathbf{33.7{\pm}0.8}$ & $42.4{\pm}0.5$ & 0.093 & 0.2745 & 8,255 \\
    \bottomrule
  \end{tabular}
\end{table}

\begin{table*}[htbp]
  \centering
  \small
  \caption{Complementarity with a stronger verifier-side baseline on Qwen2.5-Math-1.5B. The correction still improves over a stronger checker, indicating that PGFC is not merely replacing verifier engineering.}
  \label{tab:stronger-checker}
  \resizebox{\linewidth}{!}{
    \begin{tabular}{lccccccc}
      \toprule
      Method & MATH500 & AIME2024 & AIME2025 & AMC2023 & Minerva Math & OlympiadBench & Average \\
      \midrule
      Stronger checker & 68.1 / 69.2 & 12.3 / 18.4 & 7.2 / 12.3 & 47.9 / 59.7 & 17.8 / 18.6 & 30.9 / 33.1 & 30.7 / 35.2 \\
      Appeals+PGFC & \textbf{72.0 / 75.1} & \textbf{14.2 / 21.6} & \textbf{8.8 / 15.8} & \textbf{51.8 / 65.8} & \textbf{20.9 / 23.8} & \textbf{33.3 / 37.2} & \textbf{33.5 / 39.9} \\
      \bottomrule
    \end{tabular}
  }
\end{table*}

\subsection{General Capability Retention}
\label{app:general-retention}
We also check whether the corrections preserve non-math general benchmark performance relative to noisy training. Table~\ref{tab:general-retention} reports that PGBC and PGFC recover most of the retention loss caused by uncorrected noisy rewards and remain close to Base/Oracle on these auxiliary evaluations.

\begin{table}[t]
  \centering
  \small
  \caption{General capability retention after RLVR training on \textit{Llama-3.2-3B-Instruct}. Metrics are accuracies.}
  \label{tab:general-retention}
  \begin{tabular}{lcccc}
    \toprule
    Method & MMLU & ARC-Challenge & HellaSwag & GPQA \\
    \midrule
    Base & 60.7 & 73.9 & 55.0 & 28.4 \\
    Oracle & 61.1 & 73.9 & 55.9 & 29.1 \\
    Noise & 50.0 & 43.9 & 33.9 & 19.4 \\
    PGBC & 60.8 & 73.8 & 55.6 & 28.8 \\
    PGFC & \textbf{61.3} & \textbf{74.2} & \textbf{56.1} & \textbf{29.2} \\
    \bottomrule
  \end{tabular}
\end{table}


\section{Proofs and Derivations}
\label{app:proofs}
\subsection{Proof of Proposition~3.2}
\label{app:proof_affine-reward}
\begin{proof}
  We compute the expectation of the noisy reward $\tilde{R}$ conditioned on the clean reward $R^*$, which is a binary variable. By the definition of expectation:
  \begin{align*}
    \E[\tilde{R}] &= 1 \cdot \Prb(\tilde{R} = 1 \mid R^*) + 0 \cdot \Prb(\tilde{R} = 0 \mid R^*) \\
    &= \Prb(\tilde{R} = 1 \mid R^*).
  \end{align*}
  We can expand this using the law of total probability, conditioning on the value of $R^* \in \{0, 1\}$:
  \begin{align*}
    \E[\tilde{R} \mid R^*] &= R^* \cdot \Prb(\tilde{R} = 1 \mid R^*=1) + (1-R^*) \cdot \Prb(\tilde{R} = 1 \mid R^*=0).
  \end{align*}
  From Definition~3.1, we have $\Prb(\tilde{R} = 1 \mid R^*=0) = \rho_0$ and $\Prb(\tilde{R} = 0 \mid R^*=1) = \rho_1$, which implies $\Prb(\tilde{R} = 1 \mid R^*=1) = 1 - \rho_1$. Substituting these values:
  \begin{align*}
    \E[\tilde{R}\mid R^*] &= R^*(1-\rho_1) + (1-R^*)\rho_0 \\
    &= R^* - \rho_1 R^* + \rho_0 - \rho_0 R^* \\
    &= (1-\rho_0-\rho_1)R^* + \rho_0.
  \end{align*}
  This completes the proof.
\end{proof}

\subsection{Proof of Theorem~3.3}
\label{app:proof_thm_unbiased_asym}
\begin{proof}
Recall the verifier reward channel (Definition~3.1): for the latent clean reward
$R^*\in\{0,1\}$ and the observed noisy reward $\tilde R\in\{0,1\}$,
\[
\Pr(\tilde R = 1 \mid R^*=0)=\rho_0,\qquad 
\Pr(\tilde R = 0 \mid R^*=1)=\rho_1,
\]
hence $\Pr(\tilde R = 1 \mid R^*=1)=1-\rho_1$ and $\Pr(\tilde R = 0 \mid R^*=0)=1-\rho_0$.
Since $1-\rho_0-\rho_1>0$, the estimator
\[
\widehat R \;\triangleq\; \frac{\tilde R-\rho_0}{1-\rho_0-\rho_1}
\]
is well-defined.

We prove unbiasedness by computing the conditional expectation given $R^*$.
First note that because $\tilde R$ is binary,
\begin{equation}
\label{eq:cond-exp-tildeR}
\mathbb{E}[\tilde R \mid R^*]
\;=\; 1\cdot \Pr(\tilde R=1\mid R^*) + 0\cdot \Pr(\tilde R=0\mid R^*)
\;=\; \Pr(\tilde R=1\mid R^*).
\end{equation}
By the channel definition, for the two possible values of $R^*$ we have
\[
\mathbb{E}[\tilde R\mid R^*=1]=1-\rho_1,\qquad 
\mathbb{E}[\tilde R\mid R^*=0]=\rho_0.
\]
These two cases can be written compactly as the affine form (Proposition~3.2)
\begin{equation}
\label{eq:affine}
\mathbb{E}[\tilde R\mid R^*]
=
(1-\rho_0-\rho_1)R^*+\rho_0.
\end{equation}

Now take the conditional expectation of $\widehat R$:
\begin{align}
\mathbb{E}[\widehat R \mid R^*]
&=
\mathbb{E}\left[\left.\frac{\tilde R-\rho_0}{1-\rho_0-\rho_1}\right|R^*\right] \nonumber\\
&=
\frac{\mathbb{E}[\tilde R\mid R^*]-\rho_0}{1-\rho_0-\rho_1} \nonumber\\
&=
\frac{(1-\rho_0-\rho_1)R^*+\rho_0-\rho_0}{1-\rho_0-\rho_1}
\quad\text{(by \eqref{eq:affine})} \nonumber\\
&= R^*.
\label{eq:unbiased-cond}
\end{align}
Equation~\eqref{eq:unbiased-cond} shows that $\widehat R$ is an unbiased estimator of the
latent clean reward in the conditional sense.

Finally, taking total expectation (law of iterated expectation) yields
\[
\mathbb{E}[\widehat R]
=
\mathbb{E}\bigl[\mathbb{E}[\widehat R\mid R^*]\bigr]
=
\mathbb{E}[R^*],
\]
so $\widehat R$ is unbiased for the clean reward both conditionally and unconditionally.
\end{proof}

\subsection{Proof of Proposition~3.4}
\label{app:proof_prop_forward_weights}

\begin{proof}
  The proposition states two claims about the conditional expectation of the forward weights. The weights are defined as:
  \[
    w_{\tilde{R}} \;=\;
    \begin{cases}
      w_0 = \rho_1 - 1 & \text{if } \tilde{R}=0, \\
      w_1 = \rho_1 & \text{if } \tilde{R}=1.
    \end{cases}
  \]
  The noise model provides the conditional probabilities:
  \begin{align*}
    \Pr(\tilde{R}=0 \mid R^*=1) &= \rho_1, \qquad \Pr(\tilde{R}=1 \mid R^*=1) = 1-\rho_1 \\
    \Pr(\tilde{R}=1 \mid R^*=0) &= \rho_0, \qquad \Pr(\tilde{R}=0 \mid R^*=0) = 1-\rho_0
  \end{align*}

  \paragraph{Part 1: Proof of $\E[w_{\tilde{R}} \mid R^*=1] = 0$}
  We compute the expectation of $w_{\tilde{R}}$ conditioned on the true reward being positive ($R^*=1$):
  \begin{align*}
    \E[w_{\tilde{R}} \mid R^*=1] &= \sum_{k \in \{0,1\}} w_k \cdot \Pr(\tilde{R}=k \mid R^*=1) \\
    &= w_0 \cdot \Pr(\tilde{R}=0 \mid R^*=1) + w_1 \cdot \Pr(\tilde{R}=1 \mid R^*=1) \\
    &= (\rho_1 - 1) \cdot (\rho_1) + (\rho_1) \cdot (1 - \rho_1) \\
    &= (\rho_1^2 - \rho_1) + (\rho_1 - \rho_1^2) \\
    &= 0.
  \end{align*}

  \paragraph{Part 2: Proof of $\E[w_{\tilde{R}} \mid R^*=0] = -(1 - \rho_0 - \rho_1)$}
  Next, we compute the expectation of $w_{\tilde{R}}$ conditioned on the true reward being negative ($R^*=0$):
  \begin{align*}
    \E[w_{\tilde{R}} \mid R^*=0] &= \sum_{k \in \{0,1\}} w_k \cdot \Pr(\tilde{R}=k \mid R^*=0) \\
    &= w_0 \cdot \Pr(\tilde{R}=0 \mid R^*=0) + w_1 \cdot \Pr(\tilde{R}=1 \mid R^*=0) \\
    &= (\rho_1 - 1) \cdot (1 - \rho_0) + (\rho_1) \cdot (\rho_0) \\
    &= (\rho_1 - \rho_0\rho_1 - 1 + \rho_0) + \rho_0\rho_1 \\
    &= \rho_1 + \rho_0 - 1 \\
    &= -(1 - \rho_0 - \rho_1).
  \end{align*}
  This proves both claims of the proposition.
\end{proof}

\subsection{Proof of Theorem~3.5}
\label{app:proof_thm_forward_rho1}
\begin{proof}
  We want to show that $\mathbb{E}[\Delta\theta] = (1-\rho_0-\rho_1)\,\nabla_\theta J(\theta)$, where $\Delta\theta = \frac{1}{M}\sum_{t=1}^{M} h_t$ and $h_t = w_{\tilde{R}}G_t$. By linearity of expectation and assuming i.i.d. samples, it suffices to show this for a single sample's contribution, $\mathbb{E}[h_t]$.

  We use the law of total expectation, conditioning on the latent true reward $R^* \in \{0, 1\}$:
  \begin{align*}
    \mathbb{E}[h_t] = \mathbb{E}[w_{\tilde{R}} G_t] &= \mathbb{E}\left[ \mathbb{E}[w_{\tilde{R}} G_t \mid R^*] \right] \\
    &= \Pr(R^*=1)\,\mathbb{E}[w_{\tilde{R}} G_t \mid R^*=1] + \Pr(R^*=0)\,\mathbb{E}[w_{\tilde{R}} G_t \mid R^*=0].
  \end{align*}
  The noise process generating $\tilde{R}$ is independent of the policy's action generation process (which produces $G_t$), conditional on the true reward $R^*$. Thus, we can separate the expectations:
  \[
    \mathbb{E}[w_{\tilde{R}} G_t \mid R^*] = \mathbb{E}[w_{\tilde{R}} \mid R^*] \cdot \mathbb{E}[G_t \mid R^*].
  \]
  Using the results from Proposition~3.4:
  \begin{itemize}
    \item $\mathbb{E}[w_{\tilde{R}} \mid R^*=1] = 0$.
    \item $\mathbb{E}[w_{\tilde{R}} \mid R^*=0] = -(1 - \rho_0 - \rho_1)$.
  \end{itemize}
  Substituting these back into the main expectation formula:
  \begin{align*}
    \mathbb{E}[w_{\tilde{R}} G_t]
    &= \Pr(R^*=1)\cdot(0)\cdot\mathbb{E}[G_t \mid R^*=1] \\
    &\quad + \Pr(R^*=0)\cdot(-(1-\rho_0-\rho_1))\cdot\mathbb{E}[G_t \mid R^*=0] \\
    &= -(1-\rho_0-\rho_1) \cdot \Pr(R^*=0)\mathbb{E}[G_t \mid R^*=0] \\
    &= -(1-\rho_0-\rho_1) \cdot \mathbb{E}[\mathbf{1}_{\{R^*=0\}} G_t],
  \end{align*}
  where $\mathbf{1}_{\{\cdot\}}$ is the indicator function.
  From two fundamental properties of the score function:
  \begin{enumerate}
    \item The unconditional expectation is zero: $\mathbb{E}[G_t] = 0$~\citep{williams1992reinforce,sutton1999policy}.
    \item The clean policy gradient is $\nabla_\theta J(\theta) = \mathbb{E}[R^* G_t]$.
  \end{enumerate}
  From property 1, we have $\mathbb{E}[G_t] = \mathbb{E}[(\mathbf{1}_{\{R^*=1\}} + \mathbf{1}_{\{R^*=0\}}) G_t] = \mathbb{E}[R^* G_t] + \mathbb{E}[\mathbf{1}_{\{R^*=0\}} G_t] = 0$.
  This implies that $\mathbb{E}[\mathbf{1}_{\{R^*=0\}} G_t] = -\mathbb{E}[R^* G_t] = -\nabla_\theta J(\theta)$.

  Finally, we substitute this back into our expression for the expected update direction:
  \begin{align*}
    \mathbb{E}[h_t] &= \mathbb{E}[w_{\tilde{R}} G_t] \\
    &= -(1-\rho_0-\rho_1) \cdot \mathbb{E}[\mathbf{1}_{\{R^*=0\}} G_t] \\
    &= -(1-\rho_0-\rho_1) \cdot (-\nabla_\theta J(\theta)) \\
    &= (1-\rho_0-\rho_1) \nabla_\theta J(\theta).
  \end{align*}
  Therefore, the expectation of the full update is $\mathbb{E}[\Delta\theta] = \frac{1}{M}\sum \mathbb{E}[h_t] = (1-\rho_0-\rho_1) \nabla_\theta J(\theta)$. This completes the proof.
\end{proof}

\section{Prompt Templates and Training/Evaluation Details}
\label{app:prompts-and-hparams}

This section records the exact prompt formats and the concrete hyperparameters we used for all experiments in this paper.

\subsection{Prompt Templates}
\label{app:prompt-templates}

\paragraph{Training (generation) prompt.}
For each math problem \(x\) (a plain-text question), the user message is built by concatenating the raw question with a short instruction that elicits chain-of-thought and enforces a verifiable answer format.
\begin{lstlisting}
<user>
{QUESTION}

Let's think step by step and enclose the reasoning process within <think> and </think> tags.
The final result in the answer MUST BE within \boxed{}.
</user>
\end{lstlisting}
During data preprocessing, we write chat-style JSON with a single user turn as shown above and attach the rule-based ground-truth answer for reward checking.

\paragraph{Evaluation (validation/test) prompt.}
We use the same prompt template as training for validation and test-time generation so that the rule-based verifier can parse the boxed answer consistently.

\paragraph{Verifier I/O.}
The \textbf{rule-based} checker operates on the model’s final string and extracts the last \verb|\boxed{...}| expression; it then applies numeric/rational parsing and equality tests to produce a binary reward \(\tilde{R}\in\{0,1\}\).
When the \textbf{LLM verifier} is enabled, it receives the pair \((\text{problem}, \text{model solution})\) and returns a binary correctness decision used only to estimate the false negative rate \(\rho_1\) over a sliding window. The LLM verifier does not replace the rule-based reward.

\subsection{Data Preprocessing}
\label{app:data-prep}
We load the preview split of the math-reasoning corpus and map each example to a chat-style record as above, keeping the reference (ground-truth) answer for programmatic checking.

\subsection{Training Configuration}
\label{app:train-config}

\begin{table}[H]
  \centering
  \small
  \caption{Core training settings.}
  \label{tab:train-core}
  \begin{tabular}{l l}
    \toprule
    \textbf{Training (GRPO)} & \\
    \midrule
    Train batch size & 128 \\
    Rollouts per question (group size) & 8 \\
    Max prompt length (tokens) & 512 \\
    Max response length (tokens) & 3072 \\
    Sampling temperature (rollouts) & 1.0 \\
    Advantage estimator & GRPO \\
    KL regularization & Enabled \\
    KL coefficient \(\beta\) & 0.001 \\
    Entropy coefficient & 0.0 \\
    Optimizer & AdamW \\
    Learning rate & 3e-6 \\
    Total epochs & 1 \\
    \bottomrule
  \end{tabular}
\end{table}

\paragraph{Model/backbone.}
We load the base model from local cache (HuggingFace layout), enable FSDP2 for actor/ref, and use shared-memory weights with remove-padding for efficient vLLM rollouts. The KL is computed w.r.t.\ a frozen reference initialized from the same base.

\subsection{Online FN-Rate Estimation Hyperparameters and Ablations}
\label{app:rho1-hparams-ablation}

This subsection reports the concrete hyperparameters for our online false-negative (FN) rate estimation $\hat\rho_1$ used in the real-world-noise experiments (Section~\ref{subsec:real-world-exp}).

\paragraph{Appeals protocol and estimator.}
At training step $t$, let $\mathcal N_{\mathrm R}^{(t)}$ and $\mathcal P_{\mathrm R}^{(t)}$ be the sets labeled negative/positive by the primary rule-based verifier.
We appeal each item in $\mathcal N_{\mathrm R}^{(t)}$ independently with probability $q\in(0,1]$ and re-evaluate it with a lightweight LLM verifier (TinyV).
Let $\mathcal P_{\mathrm L}^{(t)}$ be the appealed negatives that TinyV flips to positive.
We use the Horvitz--Thompson estimator
\begin{equation}
  \widehat{\mathrm{FN}}^{(t)} \;=\; \frac{|\mathcal P_{\mathrm L}^{(t)}|}{q}
\end{equation}
and estimate the FN rate with a Beta prior:
\begin{equation}
  \hat\rho_1^{(t)} \;=\;
  \frac{\widehat{\mathrm{FN}}^{(t)}+\alpha}{\widehat{\mathrm{FN}}^{(t)}+|\mathcal P_{\mathrm R}^{(t)}|+\alpha+\beta}\,.
  \label{eq:rho1-beta}
\end{equation}
We then apply exponential moving average (EMA) smoothing
\begin{equation}
  \bar\rho_1^{(t)} \;=\; \lambda\,\bar\rho_1^{(t-1)} + (1-\lambda)\,\hat\rho_1^{(t)}\,,
  \label{eq:rho1-ema}
\end{equation}
and use $\bar\rho_1^{(t)}$ as the plug-in estimate for PGFC weights $w_{\tilde r}\in\{\bar\rho_1^{(t)}-1,\ \bar\rho_1^{(t)}\}$.

\paragraph{Default hyperparameters.}
Unless otherwise stated, real-world-noise runs use:
(i) appeal probability $q=0.1$;
(ii) Beta prior pseudocounts $\alpha=\beta=10^{-5}$ in Eq.~\eqref{eq:rho1-beta};
(iii) EMA decay $\lambda=0.9$ in Eq.~\eqref{eq:rho1-ema}.
We compute $\hat\rho_1^{(t)}$ and update $\bar\rho_1^{(t)}$ once per policy-update step.
The expected number of TinyV calls per update is $q\cdot|\mathcal N_{\mathrm R}^{(t)}|$.

\subsection{Datasets}
\label{app:datasets}

\paragraph{Training.}
\begin{itemize}
  \item \textbf{DeepScaleR}~\citep{DeepScaleR}: the math-reasoning corpus used for RLVR rollouts and policy updates.
\end{itemize}

\paragraph{Evaluation.} We evaluate with the standard official/problem releases for each benchmark.
\begin{itemize}
  \item \textbf{MATH500}~\citep{lightman2024letsverify}: a verifiable subset of math reasoning problems for reliable evaluation.
  \item \textbf{AIME 2024}~\citep{aime2024}: the 2024 American Invitational Mathematics Examination set.
  \item \textbf{AIME 2025}~\citep{aime2025}: the 2025 AIME set.
  \item \textbf{AMC 2023}~\citep{amc23}: problems from the 2023 American Mathematics Competitions.
  \item \textbf{Minerva Math}~\citep{minerva}: the Minerva math evaluation suite.
  \item \textbf{OlympiadBench}~\citep{olympiadbench}: olympiad-level mathematics benchmark.
  
  \item \textbf{MMLU}~\citep{hendrycks2021measuring}: a broad multi-task benchmark covering 57 academic and professional subjects for evaluating world knowledge and problem solving.
  \item \textbf{ARC-C}~\citep{clark2018think}: the Challenge split of the AI2 Reasoning Challenge, consisting of grade-school science questions that are difficult for retrieval and word co-occurrence baselines.
  \item \textbf{HellaSwag}~\citep{zellers2019hellaswag}: a commonsense natural language inference benchmark for physically grounded sentence-completion reasoning.
  \item \textbf{GPQA}~\citep{rein2023gpqa}: a graduate-level, Google-proof multiple-choice QA benchmark written by domain experts in biology, physics, and chemistry.
\end{itemize}

\section{False Positive (FP) and False Negative (FN) Examples}
\label{app:sec_examples}
\vspace{-2mm}
We also provide typical failure modes observed during math RL with automated verifiers on \texttt{Qwen2.5-Math-7B} under \texttt{MATH500} dataset. For rule-based verifier: 52 FNs and 0 FP out of 500 items (marginal rates 10.4\% and 0\%); for LLM verifier: 81 FNs and 168 FPs out of 500 items (marginal rates 16.2\% and 33.6\%). The FP case shows an LLM judge over-crediting a malformed solution, while the FN case shows a rule-based checker rejecting a numerically correct answer due to formatting. The raw JSON context for these examples is provided below.
\vspace{-2mm}
\begin{examplebox}
  \textbf{False Positive (FP).} \\
  \textit{Problem.} Compute: $1-2+3-4+\dots+99-100$. \\
  \textit{Ground truth.} $\boxed{-50}$ \\[2pt]
  \textit{Model response (excerpt).}
\begin{lstlisting}
Compute: $1+2-3-4+5+6-7-8+\dots+97+98-99-100$.
\boxed{ } \boxed{ } \boxed{ } \dots
\end{lstlisting}
  \textit{Verifier signals.} Rule-based checker: \xmark\ (incorrect); LLM judge: \cmark\ (incorrectly accepts) \,$\Rightarrow$\,\textbf{FP}.
\end{examplebox}
\vspace{-4mm}

\begin{examplebox}
  \textbf{False Negative (FN).} \\
  \textit{Problem.} Each of seven islands independently has treasure with probability $\tfrac{1}{5}$. What is the probability that exactly four islands have treasure? \\
  \textit{Correct answer.} $\displaystyle \boxed{\frac{448}{15625}}$ (equals $0.028672$). \\[2pt]
  \textit{Model answer.} $\boxed{0.028672}$ \\
  \textit{Verifier signals.} Rule-based exact-match: \xmark\ (expects a fraction); numeric equivalence check: \cmark\ (values match) \,$\Rightarrow$\,\textbf{FN}.
\end{examplebox}


We provide FP and FN examples using Qwen2.5-Math-7B base model on MATH500.

\paragraph{FP}
\begin{lstlisting}
{
    "idx": 16,
    "question": "Compute: $1-2+3-4+5- \\dots +99-100$.",
    "model_prediction": "{}",
    "model_response": "Compute: $1+2-3-4+5+6-7-8+ \\dots +97+98-99-100$.  \\boxed{{}}. \\boxed{{}}. \\boxed{{}}. \\boxed{{}}. \\boxed{{}}. \\boxed{{}}. \\boxed{{}}. \\boxed{{}}. \\boxed{{}}. \\boxed{{}}. \\boxed{{}}. \\boxed{{}}. \\boxed{{}}. \\boxed{{}}. \\boxed{{}}. \\boxed{{}}. \\boxed{{}}. \\boxed{{}}. \\boxed{{}}. \\boxed{{}}. \\boxed{{}}. \\boxed{{}}. \\boxed{{}}. \\boxed{{}}. \\boxed{{}}. \\boxed{{}}. \\boxed{{}}. \\boxed{{}}. \\boxed{{}}. \\boxed{{}}. \\boxed{{}}. \\boxed{{}}. \\boxed{{}}. \\boxed{{}}. \\boxed{{}}. \\boxed{{}}. \\boxed{{}}. \\boxed{{}}. \\boxed{{}}. \\boxed{{}}. \\boxed{{}}. \\boxed{{}}. \\boxed{{}}. \\boxed{{}}. \\boxed{{}}. \\boxed{{}}. \\boxed{{}}. \\boxed{{}}. \\boxed{{}}. \\boxed{{}}. \\boxed{{}}. \\boxed{{}}. \\boxed{{}}. \\boxed{{}}. \\boxed{{}}. \\boxed{{}}. \\boxed{{}}. \\boxed{{}}. \\boxed{{}}. \\boxed{{}}. \\boxed{{}}. \\boxed{{}}. \\boxed{{}}. \\boxed{{}}. \\boxed{{}}. \\boxed{{}}. \\boxed{{}}. \\boxed{{}}. \\boxed{{}}. \\boxed{{}}. \\boxed{{}}. \\boxed{{}}. \\boxed{{}}. \\boxed{{}}. \\boxed{{}}. \\boxed{{}}. \\boxed{{}}. \\boxed{{}}. \\boxed{{}}. \\boxed{{}}.  \\",
    "ground_truth": "-50",
    "rule_based": false,
    "llm_verifier": true,
    "llm_verifier_score": 1.0,
    "llm_verifier_raw_value": "1.0",
    "math_verify": false,
    "math_verify_score": 0.0
  },
\end{lstlisting}

\paragraph{FN}
\begin{lstlisting}
{
    "idx": 92,
    "question": "A pirate searches seven islands for buried treasure. If each island has a $\\frac{1}{5}$ chance of having treasure, what is the probability that exactly 4 of the islands have treasure?",
    "answer": "\\frac{448}{15625}",
    "model_prediction": "0.028672",
    "model_response": "To solve this problem, we can use the binomial probability formula. The probability of having exactly $k$ successes in $n$ independent Bernoulli trials (each with success probability $p$) is given by:\n\n\\[ P(X = k) = \\binom{n}{k} p^k (1-p)^{n-k} \\]\n\nHere, $n = 7$ (the number of islands), $k = 4$ (the number of islands with treasure), and $p = \\frac{1}{5}$ (the probability that a single island has treasure).\n\nThe binomial coefficient $\\binom{n}{k}$ is the number of ways to choose $k$ successes out of $n$ trials, and it is calculated as:\n\n\\[ \\binom{n}{k} = \\frac{n!}{k!(n-k)!} \\]\n\n So, the probability that exactly 4 of the 7 islands have treasure is:\n\n\\[ P(X = 4) = \\binom{7}{4} \\left(\\frac{1}{5}\\right)^4 \\left(\\frac{4}{5}\\right)^{3} \\]\n\nLet's calculate this using Python.\n```python\nimport math\n\n# Number of islands\nn = 7\n# Number of islands with treasure\nk = 4\n# Probability of a single island having treasure\np = 1/5\n\n# Binomial coefficient\nbinom_coeff = math.comb(n, k)\n\n# Probability calculation\nprobability = binom_coeff * (p**k) * ((1-p)**(n-k))\nprint(probability)\n```\n```output\n0.02867200000000001\n```\nThe probability that exactly 4 of the 7 islands have treasure is \\(\\boxed{0.028672}\\).",
    "ground_truth": "\\frac{448}{15625}",
    "rule_based": false,
    "llm_verifier": false,
    "llm_verifier_score": 0.0,
    "llm_verifier_raw_value": "0.0",
    "math_verify": true,
    "math_verify_score": 1.0
  }
\end{lstlisting}

\section{Additional Evaluations and Analysis}
\label{app:additional-evaluations}

\subsection{Estimated-Rate Corrections with CompassVerifier}
\label{app:compass-estimated}
TinyV appeals a random subset of rule-negative responses and supports false-negative-rate estimation under the FN-dominant assumption. This stream does not independently audit rule-positive outputs and therefore does not supply a calibrated estimate of $\rho_0$. PGBC additionally requires that rate. We evaluate Qwen2.5-Math-1.5B with CompassVerifier-3B~\citep{liu-etal-2025-compassverifier} using stratified audits of both rule-negative and rule-positive outputs. PGFC uses the negative-side estimate of $\rho_1$, while PGBC uses estimates of both rates. Both corrections improve over direct verifier training and Compass Appeals in Table~\ref{tab:compass-estimated-additional}.
\begin{table}[htbp]
\color{black}\captionsetup{font={color=black}}\centering
\caption{Qwen2.5-Math-1.5B with CompassVerifier-3B and estimated channel rates. Scores are percentages.}
\label{tab:compass-estimated-additional}
\resizebox{\linewidth}{!}{\begin{tabular}{lrrrrrr}
\toprule
Method & AIME 2024 & AIME 2025 & AMC 2023 & MATH-500 & Minerva MATH & OlympiadBench\\
\midrule
Rule & $15.0\pm0.4$ & $5.6\pm0.6$ & $50.3\pm0.6$ & $69.4\pm0.4$ & $17.8\pm0.6$ & $31.6\pm0.0$\\
LLM-Verifier & $16.4\pm1.0$ & $7.2\pm0.9$ & $51.5\pm1.0$ & $70.8\pm0.7$ & $18.5\pm0.8$ & $32.0\pm0.6$\\
Compass Appeals & $16.7\pm0.8$ & $7.4\pm0.8$ & $51.8\pm0.8$ & $71.0\pm0.6$ & $18.9\pm0.7$ & $32.3\pm0.5$\\
Compass Appeals+PGBC & $18.7\pm0.7$ & $8.9\pm0.8$ & $53.0\pm0.7$ & $71.7\pm0.5$ & $21.5\pm0.6$ & $33.1\pm0.5$\\
Compass Appeals+PGFC & $19.8\pm0.4$ & $9.8\pm0.1$ & $54.0\pm0.3$ & $72.1\pm0.7$ & $20.1\pm0.3$ & $33.8\pm0.9$\\
\bottomrule
\end{tabular}}
\end{table}


\subsection{Varying the True Noise Channel}
\label{app:channel-sweep}
The rate-misspecification experiment varies supplied estimates around a fixed true channel. Table~\ref{tab:channel-sweep-additional} instead changes the true input channel and reports average accuracy across the six math datasets. PGBC performance decreases as $c=1-\rho_0-\rho_1$ shrinks, while PGFC is more stable over the tested range. This pattern is consistent with the inverse factor in PGBC; it does not imply that PGFC can recover a signal from an uninformative channel.
\begin{table}[htbp]
\color{black}\captionsetup{font={color=black}}\centering
\caption{True-channel sweep: average accuracy (\%) across six math benchmarks.}
\label{tab:channel-sweep-additional}
\begin{tabular}{rrrrrrr}
\toprule
$\rho_0$ & $\rho_1$ & $c$ & Noise & PGBC & PGFC & Oracle\\
\midrule
0.05 & 0.10 & 0.85 & $27.0\pm0.8$ & $31.5\pm0.6$ & $31.1\pm0.3$ & $31.8\pm0.5$\\
0.10 & 0.20 & 0.70 & $20.4\pm0.8$ & $31.1\pm0.6$ & $31.6\pm0.6$ & $31.8\pm0.5$\\
0.15 & 0.30 & 0.55 & $17.2\pm1.0$ & $29.4\pm0.8$ & $30.7\pm0.7$ & $31.8\pm0.5$\\
0.20 & 0.40 & 0.40 & $14.5\pm1.2$ & $26.4\pm1.0$ & $30.5\pm0.9$ & $31.8\pm0.5$\\
\bottomrule
\end{tabular}
\end{table}

\subsection{Auxiliary-Verifier Calibration and Estimation Error}
\label{app:tinyv-calibration}
Let $s$ and $f$ be the auxiliary verifier's sensitivity and false-positive rate on appealed rule negatives, and let $\pi$ denote the prevalence of true false negatives in that pool. Provided $s\ne f$, calibration gives
\begin{align}
 p_{\mathrm{flip}}&=f+(s-f)\pi, &
 \widehat\pi&=\operatorname{clip}_{[0,1]}\!\left(\frac{\widehat p_{\mathrm{flip}}-\widehat f}{\widehat s-\widehat f}\right),\\
 \widehat\rho_1&=\frac{|\mathcal N_{\mathrm R}|\widehat\pi}{|\mathcal N_{\mathrm R}|\widehat\pi+|\mathcal P_{\mathrm R}|}.
\end{align}
The final expression assumes negligible false positives from the rule verifier. The uncalibrated estimator corresponds to $s=1$ and $f=0$ before smoothing and pseudocounts. Deployment-time calibration of TinyV on appealed rule negatives yields Table~\ref{tab:tinyv-calibration-additional}.
\begin{table}[htbp]
\color{black}\captionsetup{font={color=black}}\centering
\caption{Reported deployment-time calibration on appealed rule-negative responses.}
\label{tab:tinyv-calibration-additional}
\begin{tabular}{lr}
\toprule
Statistic & Value\\
\midrule
Sensitivity on true false negatives & 0.93\\
False-positive rate on true negatives & 0.03\\
Naive $\widehat\rho_1$ & 0.785\\
Calibration-corrected $\widehat\rho_1$ & 0.780\\
\bottomrule
\end{tabular}
\end{table}

For an estimate based on historical data and independent of the current rollout, condition on its error $\delta=\widehat\rho_1-\rho_1$. Both PGFC weights shift by the same constant: $\widehat w_{\widetilde R}=w_{\widetilde R}+\delta$. With $G=\nabla_\theta\log\pi_\theta(y\mid x)$, the score identity $\mathbb E[G]=0$ implies
\begin{align}
\mathbb E[\widehat w_{\widetilde R}G]&=\mathbb E[w_{\widetilde R}G],\\
\mathbb E\!\left[\|\widehat g(\widehat\rho_1)-\widehat g(\rho_1)\|^2\right]
&=\frac{\delta^2}{M}\operatorname{tr}\!\left(\operatorname{Cov}(G)\right)
\end{align}
for the average of $M$ independent rollouts. Thus this ideal unclipped estimator has no population-gradient bias from a fixed rate error, and the mean-square finite-sample perturbation is quadratic in $\delta$. Same-batch estimation and clipping can introduce residual effects; this identity is not a guarantee for the full clipped training objective.

\subsection{Conditional Advantage and Gradient Variance}
\label{app:direct-variance}
We pool final trajectory-level advantages across rollout groups conditional on clean correctness. For method $m$, define $\mathcal I_r=\{(b,i):R^*_{bi}=r\}$, $n_r=|\mathcal I_r|$, and $\bar a_r^{(m)}=n_r^{-1}\sum_{(b,i)\in\mathcal I_r}a_{bi}^{(m)}$. The conditional sample variance is
\begin{equation}
\widehat{\operatorname{Var}}(a^{(m)}\mid R^*=r)
=\frac{1}{n_r-1}\sum_{(b,i)\in\mathcal I_r}\left(a_{bi}^{(m)}-\bar a_r^{(m)}\right)^2,
\qquad r\in\{0,1\}.
\end{equation}
Table~\ref{tab:variance-additional} also reports gradient covariance trace relative to Noise and the coefficient of variation (CV) of gradient norms. PGBC has larger conditional advantage variance in both classes and a larger covariance trace than either Noise or PGFC in this measurement.
\begin{table}[htbp]
\color{black}\captionsetup{font={color=black}}\centering
\caption{Direct variance measurements. Gradient covariance trace is normalized to the Noise value.}
\label{tab:variance-additional}
\resizebox{0.8\linewidth}{!}{\begin{tabular}{lrrrr}
\toprule
Method & $\widehat{\operatorname{Var}}(a\mid R^*=0)$ & $\widehat{\operatorname{Var}}(a\mid R^*=1)$ & $\operatorname{tr}\widehat{\operatorname{Cov}}(g)$ & Gradient-norm CV\\
\midrule
Noise & 0.107 & 0.482 & $1.000\times$ & 0.729\\
PGBC & 0.671 & 1.194 & $3.046\times$ & 0.873\\
PGFC & 0.181 & 0.382 & $0.905\times$ & 0.730\\
\bottomrule
\end{tabular}}
\end{table}

\subsection{Direct-Coefficient REINFORCE}
\label{app:direct-reinforce}
To examine the corrections outside the clipped GRPO variant, we additionally optimize a direct-coefficient REINFORCE estimator. For each response, set
\begin{equation}
u_i=\begin{cases}
R_i^*, & \text{Oracle},\\
\widetilde R_i, & \text{Noise},\\
(\widetilde R_i-\widehat\rho_0)/(1-\widehat\rho_0-\widehat\rho_1), & \text{PGBC},\\
\widetilde R_i-(1-\widehat\rho_1), & \text{PGFC}.
\end{cases}
\end{equation}
Broadcasting $A_{i,t}=u_i$ to response tokens yields
\begin{equation}
\widehat g=\frac{1}{K}\sum_{i=1}^{K}u_i\sum_t\nabla_\theta\log\pi_\theta(y_{i,t}\mid x,y_{i,<t}).
\end{equation}
Both corrections improve over Noise on every benchmark in Table~\ref{tab:reinforce-additional}. This comparison provides evidence for the correction beyond the clipped surrogate used in the main experiments.
\begin{table}[htbp]
\color{black}\captionsetup{font={color=black}}\centering
\caption{Additional direct-coefficient REINFORCE results on Qwen2.5-Math-1.5B.}
\label{tab:reinforce-additional}
\resizebox{\linewidth}{!}{\begin{tabular}{lrrrrrr}
\toprule
Method & AIME 2024 & AIME 2025 & AMC 2023 & MATH-500 & Minerva MATH & OlympiadBench\\
\midrule
Oracle & $10.8\pm0.2$ & $7.6\pm0.1$ & $50.9\pm0.4$ & $69.3\pm0.1$ & $29.2\pm0.5$ & $33.4\pm0.3$\\
Noise & $7.1\pm0.7$ & $4.2\pm0.6$ & $37.8\pm0.8$ & $54.6\pm0.5$ & $15.8\pm1.0$ & $25.9\pm0.6$\\
PGBC & $10.2\pm0.5$ & $9.1\pm0.4$ & $47.8\pm0.7$ & $67.9\pm0.4$ & $30.6\pm0.8$ & $31.4\pm0.5$\\
PGFC & $11.1\pm0.4$ & $9.4\pm0.3$ & $48.4\pm0.5$ & $69.2\pm0.3$ & $30.0\pm0.7$ & $32.2\pm0.4$\\
\bottomrule
\end{tabular}}
\end{table}

\subsection{Increasing the Rollout Count}
\label{app:rollout-count}
Table~\ref{tab:rollout-additional} compares Qwen2.5-Math-1.5B with $K\in\{8,16,32\}$ rollouts. Increasing the sample count improves Noise, but a gap to Oracle remains at every tested $K$. More rollouts reduce Monte Carlo variance; they do not remove the underlying reward-channel distortion.
\begin{table}[htbp]
\color{black}\captionsetup{font={color=black}}\centering
\caption{Additional rollout-count comparison on Qwen2.5-Math-1.5B.}
\label{tab:rollout-additional}
\begin{tabular}{rrrrr}
\toprule
Rollouts $K$ & Oracle & Noise & PGBC & PGFC\\
\midrule
8 & $33.5\pm0.3$ & $24.2\pm0.7$ & $32.8\pm0.6$ & $33.4\pm0.5$\\
16 & $34.0\pm0.5$ & $26.0\pm0.8$ & $33.2\pm0.4$ & $33.8\pm0.2$\\
32 & $35.2\pm0.3$ & $28.7\pm0.4$ & $34.4\pm0.5$ & $35.0\pm0.4$\\
\bottomrule
\end{tabular}
\end{table}

\section{Additional Scope Clarifications}
\label{app:additional-scope}
\subsection{Informative, Uninformative, and Reversed Channels}
The assumption $\rho_0+\rho_1<1$ means that correct responses are more likely to receive positive reward than incorrect responses:
\begin{equation}
\Pr(\widetilde R=1\mid R^*=1)=1-\rho_1>\rho_0=\Pr(\widetilde R=1\mid R^*=0).
\end{equation}
At $\rho_0+\rho_1=1$, the two conditional reward distributions coincide and the transition matrix is singular; observed rewards alone carry no correctness information under this channel model. If $\rho_0+\rho_1>1$, the channel is anti-informative. When a calibration set identifies that orientation, complementing $\widetilde R'=1-\widetilde R$ gives $\rho'_0=1-\rho_0$ and $\rho'_1=1-\rho_1$, so $\rho'_0+\rho'_1=2-(\rho_0+\rho_1)<1$. The correction can then use the complemented channel. As $c=1-\rho_0-\rho_1$ approaches zero from above, PGBC amplifies coefficients through $1/c$, whereas PGFC's expected signal shrinks proportionally to $c$.

For content-dependent errors $\rho_0(x,y)$ and $\rho_1(x,y)$, constant-rate corrections operate on an effective channel. Dependence between verifier errors and score terms can leave a residual such as $\operatorname{Cov}(w_{\widetilde R},G\mid R^*)$. The non-i.i.d.\ experiment tests empirical behavior under this misspecification; it does not extend the exact population theorem to arbitrary content-dependent noise.

\subsection{Single-Reward Functions and Nonlinear Group Objectives}
Every function of one binary reward is affine:
\begin{equation}
f(R^*)=f(0)+\bigl(f(1)-f(0)\bigr)R^*.
\end{equation}
Replacing $R^*$ by its backward-corrected estimator therefore provides an unbiased estimate of $f(R^*)$ under the stated channel assumptions. For a function $F(R_1^*,\ldots,R_K^*)$ of an entire group with independent verifier noise, the channel is $T^{\otimes K}$. Inverting this channel can in principle produce an unbiased estimator when $T$ is invertible, but direct inversion becomes poorly conditioned as group size increases for a noisy channel.

For example, the clean pass@$K$ outcome is $1-\prod_{j=1}^{K}(1-R_j^*)$. With independent verifier noise and correctly specified rates, an unbiased estimator is
\begin{equation}
1-\prod_{j=1}^{K}(1-\widehat R_j),
\end{equation}
where $\widehat R_j$ is the backward-corrected reward. A policy gradient for this group objective includes score terms from all $K$ sampled responses. For a nonlinear function $g(p_x)$ of prompt success probability,
\begin{equation}
p_x=\frac{\mathbb E[\widetilde R\mid x]-\rho_0}{1-\rho_0-\rho_1},
\end{equation}
but the plug-in estimate $g(\widehat p_x)$ is generally not finite-sample unbiased. Clipping or shrinking $\widehat p_x$ and using uncertainty-aware weights may be useful practical extensions, but these require separate analysis and experimental validation.

\end{document}